\pdfoutput=1

\documentclass[11pt]{article}

\usepackage{latex/acl}

\usepackage{times}
\usepackage{latexsym}

\usepackage[T1]{fontenc}

\usepackage[utf8]{inputenc}

\usepackage{microtype}

\usepackage{inconsolata}
\usepackage{booktabs}
\usepackage{algorithm}
\usepackage{algpseudocode}
\usepackage{adjustbox}
\usepackage{multirow}
\usepackage{graphicx, subfigure}
\usepackage{amsmath,amsthm,amsfonts,amssymb,amscd}
\usepackage{mathtools}
\usepackage{enumitem}
\usepackage{bbm}
\usepackage{booktabs} 
\usepackage{tabularx} 
\usepackage{xcolor}
\usepackage{setspace}

\usepackage[textsize=tiny]{todonotes}
\usepackage{tcolorbox}

\title{Hard Prompts Made Interpretable: \\ Sparse Entropy Regularization for Prompt Tuning with RL}

\author{Yunseon Choi$^{1}$ \quad Sangmin Bae$^{1}$ \quad Seonghyun Ban$^{1}$ \quad Minchan Jeong$^{1}$  \\
        \vspace{2pt}
        {\bf Chuheng Zhang}$^{2}$ \quad 
        {\bf Lei Song}$^{2}$ \quad 
        {\bf Li Zhao}$^{2}$ \quad 
        {\bf Jiang Bian}$^{2}$ \quad 
        {\bf Kee-Eung Kim}$^{1}$ \\
    \vspace{2pt}
    \textsuperscript{\rm 1}KAIST AI \quad \textsuperscript{\rm 2}Microsoft Research Asia\\
    \vspace{2pt}
    \texttt{\{cys9506, kekim\}@kaist.ac.kr}}

\begin{document}
\maketitle
\begin{abstract}
With the advent of foundation models, prompt tuning has positioned itself
as an important technique for directing model behaviors and eliciting desired responses.
Prompt tuning regards selecting appropriate keywords included into the input, 
thereby adapting to the downstream task without adjusting or fine-tuning the model
parameters. 
There is a wide range of work in prompt tuning, 
from approaches that directly harness the backpropagated gradient signals from the model,
to those employing black-box optimization 
such as reinforcement learning (RL) methods.
Our primary focus is on RLPrompt, which aims to find optimal prompt tokens
leveraging soft Q-learning. While the results show promise, we have observed 
that the prompts frequently appear unnatural, which impedes their interpretability. 
We address this limitation by using sparse Tsallis entropy regularization, 
a principled approach to filtering out unlikely tokens from consideration. 
We extensively evaluate our approach across various tasks,
including few-shot text classification, unsupervised text style transfer, and textual inversion from images. 
The results indicate a notable improvement over baselines, 
highlighting the efficacy of our approach in addressing the challenges of prompt tuning.
Moreover, we show that the prompts discovered using our method are more natural and interpretable compared to those from other baselines~\cite{rlprompt}
\footnote{Code available at \url{https://github.com/Youseob/PIN}}.
\end{abstract}

\section{Introduction}
\label{sec:intro}

While the use of large-scale language models (LMs) and vision-language models (VLMs),
pre-trained on a massive amount of data, is becoming a dominant paradigm in machine learning~\cite{latent-diff,touvron2023llama,clip}, 
fine-tuning the model parameters for adaptation to downstream tasks requires a vast amount of computational time and resources. 
In this sense, \textit{prompt tuning} has emerged as a promising low-cost solution~\cite{NEURIPS2020_1457c0d6,lester-etal-2021-power}, discovering input prompts that effectively guide the pre-trained models to generate the desired outputs, while keeping the model parameters frozen.
Prompt tuning is generally categorized into two approaches, \textit{soft} and \textit{hard} prompting methods, based on their representation of prompts.


Soft prompt methods~\cite{lester-etal-2021-power,li-liang-2021-prefix} primarily focus on 
learning continuous embedding vectors at the token level, which are called as soft prompts. 
They usually perform the gradient descent to optimize these continuous embedding vectors~\cite{wu2023infoprompt,iccv-soft-prompt}. 
However, the prompts learned through soft tuning are opaque to human interpretation and 
are not compatible with other pre-trained models that do not share the same embedding spaces. 
Moreover, computing internal gradients for models is highly resource-intensive, 
especially as the number of model parameters increases, or even infeasible 
in cases where models are accessible only through APIs \cite{gpt-4}. 
These limitations necessitate an alternative approach:
discovery of the prompts composed of human-readable discrete tokens, 
referred to as \textit{hard prompts}~\cite{prasad-etal-2023-grips, shin-etal-2020-autoprompt, rlprompt}.

Hard prompts
offer numerous advantages over soft prompts: they are transferable from one pre-trained model to 
another since they are agnostic to the embedding. Moreover, they confer compositional 
control by facilitating manual merging and modification~\cite{pez}.
Despite the benefits, they require large-scale discrete optimization in principle.
Approaches for finding optimal hard prompts often employ
backpropagated gradients from models as with soft prompt methods,
while circumventing the discrete optimization by, roughly speaking, mapping the soft
prompt to the similar discrete tokens in the embedding space~\cite{pez,shin-etal-2020-autoprompt}. 
However, particularly when the model is black-box,
reinforcement learning (RL) serves as a powerful alternative tool for
optimization, as exemplified by RLPrompt~\cite{rlprompt} that uses 
soft Q-learning~\cite{sql}.


One of the key ideas behind RLPrompt is an efficient parameterization
leveraging a frozen pre-trained LM. However, as we show later in the paper,
this is accompanied by detrimental approximation error, leading to undesirable results.
In this paper, we address this limitation in a principled manner. 
Our approach leverages sparse Tsallis entropy regularization for RL~\cite{tsallis-entropy-rl}
to ignore very unlikely tokens from consideration. 
We demonstrate the effectiveness of our algorithm across various tasks, 
including few-shot text classification, unsupervised text style transfer, and 
textual inversion from images, 
comparing it against various baselines.
Most importantly, unlike hard prompts learned by baselines which are often referred to as the `secret language' of 
models due to their opacity for human interpretation, 
our learned prompts are more natural and straightforward. 
Our contributions are outlined as follows:
\vspace{-5pt}
\begin{itemize}[leftmargin=10pt]
\setlength\itemsep{-0.2em}
    \item We first identify the problem associated with the parameterization employed in RLPrompt that potentially leads to suboptimal and unnatural prompts. 
    \item We propose a principled solution to the problem using sparse Tsallis entropy~\cite{tsallis-entropy-rl}, 
    referred to as PIN (Prompts made INterpretable).
    \item We show extensive experimental results that demonstrate the effectiveness of our algorithm across various downstream tasks. 
\end{itemize}

\section{Related Works}
\label{sec:related}

\paragraph{Prompting in Language Models}
Pioneering work by \citet{NEURIPS2020_1457c0d6} 
highlighted the efficacy of using prompts for task adaptation in pre-trained language models, a technique now commonly referred to as instruction tuning. 
This approach has become standard 
in enhancing the ability of large models to execute complex, task-specific instructions.
Despite its success, 
the automated generation of effective text prompts, 
particularly hard prompts, remains a challenge. 
The work by \citet{lester-etal-2021-power} to simplify prefix tuning led to the establishment of standard soft prompt tuning, which optimizes continuous embeddings that are appended to embeddings of input tokens.
However, \citet{khashabi-etal-2022-prompt} 
pinpointed a limitation of this approach: the resulting embedding sequences often lack clear semantic interpretation.
To address these limitations,
our work focuses on the hard prompts optimization within a selected set of tokens, thereby generating task-specific and interpretable tokens.


\paragraph{Discrete Optimization for Hard Prompts}

AutoPrompt~\cite{shin-etal-2020-autoprompt} is 
an initial framework for discrete prompt optimization in transformer-based 
language models, inspiring a range of diverse methods.
These include a gradient-free phrase editing method~\citep{prasad-etal-2023-grips}, a reinforcement learning-based approach~\citep{rlprompt}, 
and an embedding optimization approach based on Langevin dynamics~\cite{shi-etal-2023-toward}. 
In our work, we first benchmark gradient-based methods like AutoPrompt~\citep{shin-etal-2020-autoprompt} and PEZ~\cite{pez} for hard prompt tuning.
AutoPrompt employs the HotFlip algorithm \citep{ebrahimi-etal-2018-hotflip} to greedily identifies optimal tokens for each position based on model gradients, while PEZ performs gradient re-projection during its continuous optimization within the embedding space. However, both methods require substantial computational cost in calculating gradients and are unsuitable for black-box models.
On the other hand, RLPrompt~\cite{rlprompt} employs a gradient-free RL-based method, serving as our primary baseline. RLPrompt introduces an efficiently parameterized network that maps continuous embedding vectors to adaptive vectors within the same space.
Despite its simplicity, RLPrompt struggles with accurately representing Q-values across all tokens, potentially leading to sub-optimal prompts as we discuss in Section~\ref{sec:issue}.


\section{Preliminaries}
\label{sec:preliminaries}
\paragraph{Hard Prompt Tuning with RL}
Hard prompt tuning is the process of discovering an optimal prompt within the token space $\mathcal{V}$, 
to efficiently tackle specific downstream tasks.
Assuming a fixed-length prompt of $L$ tokens,
this optimization can be formally defined as RL problem: 
\begin{equation}
    \max_{\boldsymbol{z} \in \mathcal{V}^L} R(\boldsymbol{y}(\boldsymbol{z}, \boldsymbol{x})).
\label{eqn:obj}
\end{equation}
Here, the objective is to find a discrete prompt $\boldsymbol{z}$
from the solution space of length-$L$ token sequences $\mathcal{V}^L$, 
which maximizes a task-specific reward function $R$ when concatenated with input $\boldsymbol{x}$.
This reward function measures the appropriateness of the model output 
$\boldsymbol{y}(\boldsymbol{z}, \boldsymbol{x})$,
i.e., the output from LMs or VLMs
when prompted with $\boldsymbol{z}$ for input $\boldsymbol{x}$.
For example, in a few-shot text classification task utilizing masked LMs as task model,
the reward function can be defined as a binary signal that indicate the correctness of model output based on available few-shot data, where model output refers to the predicted class for the [MASK] token position\footnote{The task of textual inversion from images
requires a slight deviation from the standard formulation. Instead of the reward based on the model output, it is defined as the similarity between the embedding of a target image and the text prompt.}.

Eq.~\eqref{eqn:obj} can be approached using a bandit algorithm,
which aims to identify a length-$L$ token sequence $\boldsymbol{z}$ that 
maximizes the reward $R$ without gradient information.
However, to cope with the exponentially large action space 
$\mathcal{O}(|\mathcal{V}|^L)$ for the bandit algorithm,
we can treat the optimization as a sequential decision-making process,
as in RLPrompt~\cite{rlprompt}.
More concretely,
at each time step $t$, the algorithm chooses the token $z_t$ based on the
tokens $z_{0:t-1}$ chosen at previous time steps, denoted as 
policy $\pi(z_t | z_{0:t-1})$. 
The calculation of reward $R$ is delayed until
the completion of the entire prompt sequence $\boldsymbol{z}$ to obtain the model output.
Thus, we optimize the policy $\pi$ with the reformulation of Eq.~\eqref{eqn:obj},
given by
\begin{equation*}
    \max_\pi \mathbb{E}_{\boldsymbol{z} \sim \prod_{t=0}^{L-1} \pi(z_t | z_{0:t-1})}
    [R (\boldsymbol{y}(\boldsymbol{z}, \boldsymbol{x}))] .
\end{equation*}

\paragraph{Soft Q-Learning (SQL) and RLPrompt}
RLPrompt~\cite{rlprompt} employs SQL~\citep{sql}, 
an RL algorithm that incorporates entropy regularization.
It aims to maximize the expected cumulative reward with the bonus given by
the entropy of the action distribution in order to balance the trade-off between
exploitation and exploration.
More formally, at each time step $t$, RLPrompt trains the policy with the objective:
\begin{equation}
 \max_\pi \mathbb{E}_{z\sim\pi(z|z_{0:t\text{-1}})} 
 \!\left[ Q(z_{0:t\text{-1}},z) - \alpha \log \pi(z|z_{0:t\text{-1}}) \right]\!,
 \label{eqn:sql_objective}
\end{equation}
where $Q$ is the action-value function capturing long-term reward effects as follows: 
\begin{equation*}
Q(z_{0:t\text{-1}},z) \!\triangleq\! \mathbb{E}_{z_{t\text{+1}:L\text{-1}}\sim\pi}
[R(\boldsymbol{y}(z_{0:t\text{-1}},z,z_{t\text{+1}:L\text{-1}},\boldsymbol{x})],
\end{equation*}
and $\alpha$ is the regularization coefficient.

Eq.~\eqref{eqn:sql_objective} yields an analytical solution for the optimal policy,
expressed as
\begin{equation*}
    \pi^*(z | z_{0:t-1}) = \text{softmax}\left(\frac{{Q(z_{0:t-1}, z)}}{\alpha}\right)\!.
\label{eqn:sql_policy}
\end{equation*}
However, since the action-value function $Q$ is not readily available, 
it is estimated by a neural
network parameterized by $\theta$, referred to as $Q$-network.

One of the main contributions of RLPrompt is the introduction of 
an efficient parameterization for the $Q$-network.
This involves integrating a frozen and pre-trained language model (LM), referred to as the policy LM, 
into the lower layers of the $Q$-network. 
Besides, a trainable multi-layer perceptron (MLP) layer is augmented at the upper level
for the adaptation to the downstream task.
Thus, the trainable parameter $\theta$ of the $Q$-network is only the parameters of the MLP layer.

Formally, 
given the prefix of prompt $z_{0:t-1}$,
the encoding vector $e_t \in \mathcal{E}$ obtained from the policy LM is passed through 
MLP layer $\psi_\theta$ to compute an adapted embedding $\hat{e}_t \in \mathcal{E}$,
where $\mathcal{E}$ denote the real vector space of embeddings.
Subsequently, 
this adapted embedding $\hat{e}_t$ is multiplied with the LM-head matrix of policy LM, 
$W^{\text{LM}} \in \mathbb{R}^{ |\mathcal{V}| \times \text{dim}(\mathcal{E})}$,
to get the next prompt token probabilities.
\begin{align*}
\label{eqn:rl_prompt}
    & \hat{e}_t = \psi_\theta (e_t) = \psi_\theta (\text{LM}(z_{0:t-1})) \\ 
    & Q_\theta(z_{0:t-1}, \cdot) = W^{\text{LM}}\hat{e}_t \nonumber\\
    & \pi_{\theta}(z_t | z_{0:t-1})
    := \text{softmax} \left( \frac{Q_\theta(z_{0:t-1}, z_t)}{\alpha} \right)\!, \nonumber
\end{align*}
where $W^{\text{LM}} $ is kept fixed. 
Thus, this particular parameterization makes $W^{\text{LM}}\hat{e}_t$ collectively
represent the scaled Q-values for all tokens. In other words, 
for token $z\in \mathcal{V}$ and its index $i_z$, 
its action value is calculated by $ w_{i_z}^{\top}\psi_\theta(e_t) = Q_\theta(z_{0:t-1}, z)$,
where $w_{i}$ denotes the $i$-th row vector of $W^{\text{LM}}$.

To optimize the parameter $\theta$,
the objective is to minimize the temporal difference error, 
\begin{equation}
\mathbb{E}_{\boldsymbol{z}\sim \pi_\theta}[\big( Q_\theta(z_{0:t-1}, z_{t}) - \hat{Q}(z_{0:t-1}, z_{t}) \big)^2].
\label{eqn:mse_loss}
\end{equation}
Here, the target value $\hat{Q}$ is the bootstrapped estimate of the action value, given by: 
\begin{align*}
    \hat{Q}&(z_{0:t-1}, z_{t}) \\
     & =\begin{cases}
     \gamma \alpha \log \sum_{z\in\mathcal{V}} \exp (\frac{Q_\theta(z_{0:t}, z)}{\alpha}) & \text{if }t < L-1 \\
     R (\boldsymbol{y}(\boldsymbol{z}, \boldsymbol{x})) & \text{if }t = L-1 \nonumber
     \end{cases}
     \,
\end{align*}
where $\gamma \in (0, 1]$ denotes the discount factor.

\paragraph{Sparse Tsallis Entropy Regularized Q-Learning}
Given a policy $\pi(\cdot|z_{0:t-1})$ that represents a probability distribution on 
$z$, the 
\textit{Tsallis entropy}~\cite{tsallis} with entropic index $q$ is defined as 
$S_q(\pi) = k \frac{1}{q-1} (1 - \sum_{z} \pi^q(z|z_{0:t-1}))$,
where $k$ is a positive scalar value.
It generalizes various types of entropy via the entropic index $q$.
Specifically, as $q \to 1$, 
it becomes \textit{Shannon entropy}, $S_1(\pi)=\mathbb{E}_\pi[-\log \pi(z|z_{0:t-1})]$, which is 
employed by SQL for regularizing policy. 
This results in an optimal policy that is softmax function over Q-values, 
a feature known to promote exploration within the decision-making.
However, as mentioned earlier, an intrinsic characteristic of a softmax policy is its distribution of nonzero probability mass across all actions.
On the other hand, $q=2$ yields \textit{sparse Tsallis entropy}, $S_2(\pi) = \mathbb{E}_\pi[\frac{1}{2}(1-\pi(z|z_{0:t-1}))]$, which is the focus of this paper.
Employing sparse Tsallis entropy as regularization~\cite{tsallis-entropy-rl} leads to an sparse optimal policy that concentrates probability mass on a subset set of actions.

Analogous to Eq.~\eqref{eqn:sql_objective}, 
the analytical formula of the optimal policy with the sparse Tsallis entropy regularization 
with the regularization coefficient $\alpha$ is given by
\vspace{-4pt}
\begin{align}
    \label{eq:sp_opt_policy}     
        &\pi^*(z | z_{0:t-1}) \\ 
        &= \max \left(
         \frac{Q(z_{0:t-1}, z)}{\alpha} 
        - \tau\left(\frac{Q(z_{0:t-1}, \cdot)}{\alpha}\right),
        0 \right)\!, \nonumber
\end{align}
where 
$\tau$ is the thresholding function that sets 
the action probabilities to zero when their $\alpha$-scaled action values 
fall below $\tau(Q(z_{0:t-1}, \cdot)/\alpha)$ and 
ensures the sum of these probabilities equals 1:
\vspace{-3pt}
\begin{equation*}
\tau\!\left(\frac{Q(z_{0:t-1}, \cdot)}{\alpha}\right)\!=\!\frac{\sum_{z \in S_{Q(z_{0:t-1}, \cdot)/\alpha}} 
\!\!\!\frac{Q(z_{0:t-1}, z)}{\alpha}-1}{|S_{Q(z_{0:t-1}, \cdot)/\alpha}|}, 
\end{equation*}
where the set $S_{Q(z_{0:t-1}, \cdot)/\alpha}$, referred to as the supporting set, consists of $z_{(n)}\in \mathcal{V}$
that satisfy 
\vspace{-4pt}
\begin{align*}
    & S_{Q(z_{0:t-1}, \cdot)/\alpha} =  \\
    & \{ z_{(n)} | 
    1 + n\frac{Q(z_{0:t-1}, z_{(n)})}{\alpha} 
    >\!\sum_{m=1}^{n} \frac{Q(z_{0:t-1}, z_{(m)})}{\alpha} 
     \}, \nonumber
\vspace{-4pt}
\end{align*}
with $z_{(m)}$ indicating the action with the $m$-th largest value of $Q(z_{0:t-1},z)$.
Thus, the supporting set $S_{Q(z_{0:t-1}, \cdot)/\alpha}$ 
consists of top-$K_\alpha$ tokens with the highest  action values,
and the coefficient $\alpha$ controls the cardinality $K_\alpha:=|S_{Q(z_{0:t-1}, \cdot)/\alpha}|$.
Consequently,
the sparse policy from Eq.~\eqref{eq:sp_opt_policy} 
assigns non-zero probabilities for only $K_\alpha$-actions, where
smaller $\alpha$ makes the policy sparser.

For training $Q_\theta(z_{0:t-1}, z)$, we
minimize the temporal difference error using the target value $\hat{Q}$:
\vspace{1pt}
\begin{align}
\label{eq:sp_target}
     \hat{Q}&(z_{0:t-1}, z_{t}) \\
     &= \begin{cases}
     \gamma \alpha\,  \text{\text{sparsemax}}
     \left( \frac{Q(z_{0:t}, \cdot)}{\alpha} \right) & \text{if }t < L-1 \\
     R(\boldsymbol{y}(\boldsymbol{z}, \boldsymbol{x})) & \text{if }t = L-1
     \end{cases}
     . \nonumber
\end{align}
where the sparsemax operator is defined as  
\begin{align*}
&\text{\text{sparsemax}}\left( \frac{Q(z_{0:t}, \cdot)}{\alpha} \right)  \\
&= \frac{1+\!\!\displaystyle\sum_{z\in S_{Q(z_{0:t, \cdot})/\alpha}} \textstyle
        \left( \frac{Q(z_{0:t}, z)}{\alpha} \right)^2
        - \tau \left( \frac{Q(z_{0:t}, \cdot)}{\alpha} \right)^2 }{2}, \nonumber
\end{align*}
which limits the target value estimation to top-$K_\alpha$ actions.



\section{Method}
\label{sec:method}

\subsection{Overdetermined Linear Systems} 
\label{sec:issue}
While the $Q$-network utilized by RLPrompt provides efficient parameterization, it also possesses a fundamental limitation.
Training the $Q$-network is essentially solving for an extremely overdetermined linear system, 
where approximation error is inevitable.
In order to observe this, we fix $z_{0:t-1}$ and rewrite
Eq.~\eqref{eqn:mse_loss} as the weighted least-squares problem
\begin{equation}
\label{eqn:over}
\min_{\boldsymbol{\psi} }
\| W^\text{LM} \boldsymbol{\psi} - \hat{\boldsymbol{q}} \|^2_{\pi(\cdot|z_{0:t-1})},   
\end{equation}
where $\boldsymbol{\psi} = \psi_\theta(\text{LM}(z_{0:t-1}))$ is the optimization variable,
and $\hat{\boldsymbol{q}} = \hat{Q}(z_{0:t-1},\cdot)$ is the right-hand-side vector.
The coefficient matrix, which is the LM-head matrix from the policy LM, $W^\text{LM} \in \mathbb{R}^{|\mathcal{V}|\times\dim(\mathcal{E})}$
has many more rows than columns since it is common that $|\mathcal{V}| \gg \dim(\mathcal{E})$ (e.g., $|\mathcal{V}| = 50272$ and $\dim(\mathcal{E}) = 768$
for OPT-125M), corresponding to many more constraints than variables 
resulting in inevitable approximation error. 

Together with the probabilities of tokens as weights in the least-squares formulation, 
the issue of approximation error becomes more critical. 
Tokens with high probabilities, as estimated by $\pi$, receive larger weights in the least-squares, leading to smaller approximation errors for them. 
On the other hand, the vast number of low-probability tokens are assigned with smaller weights, resulting in relatively larger approximation errors. 
Consequently, the estimated action value could become unreasonably high for these low-probability tokens, promoting the RL algorithm to excessively try out these improbable tokens.
This results in an RL approach that is overly biased towards exploration, specifically favoring the selection of insignificant low-probability tokens.
The detrimental impact of this unfortunate behavior is clearly observed by the unnatural prompts chosen by RLPrompt.

\subsection{PIN (Prompts made INterpretable)}
\label{sec:alg}

One of the straightforward solutions to mitigate the challenge of 
the overdetermined linear system is to drop the constraints that
are less important. To achieve this, we introduce the 
\textit{ignorable token set} comprised of 
tokens deemed improbable from a general language model.
The key concept is to sidestep evaluating the action values of these tokens, 
as they may adversely impact the value estimations of high-probability tokens. 
The construction of the ignorable token set entails utilizing the logit of the predictive probability
of tokens derived from the policy LM with the original embedding, and 
choosing those that score below the $k$-th largest logit.
Formally, given the prompt prefix $z_{0:t-1}$, the ignorable token set is defined by
\[
\mathcal{I}_{z_{0:t-1}} = \{ z \in \mathcal{V} \,|\, w_{i_z}^\top e_t < w_{i_{(k)}}^\top e_t \}, 
\]
where $e_t$ is the embedding vector of $z_{0:t-1}$ obtained from the policy LM,
$w_i$ is the $i$-th row vector of its LM-head matrix, and $i_{(k)}$ is the index of 
the token with $k$-th largest logit. 

Deciding the number of $k$ is important for learning effective hard prompts, 
as demonstrated by our experimental results.
If we set $k$ aggressively (i.e., small) to 
avoid constraint violation, we may end up ignoring strong candidate tokens. 
On the other hand, if we set $k$ conservatively (i.e., large), we suffer from 
the original challenge of approximation error. 
Therefore, we empirically chose $k=10000$ to ensure a sufficiently diverse set of tokens, disregarding about 80\% of the vocabulary tokens.
Nevertheless, we observed that the ignorable token set dose not work well alone
because $k$ is significantly larger than $\text{dim}(\mathcal{E})$, which is 768, even in case of OPT-125M.

Now, the remaining challenge is to address the approximation error still existent 
due to more constraints than variables for training the $Q$-network.
If we adopt SQL as the baseline RL method, 
we will end up with the same undesirable behavior due to the dense probability 
over the tokens. We thus instead employ the sparse Tsallis entropy regularized 
Q-learning described in the previous section, which yields a sparse policy 
that naturally suppresses the probability of choosing many unimportant tokens
with errors in the action value estimation.

Our algorithm, PIN (Prompts made INterpretable), is presented in Algorithm~\ref{alg:alg}. 
We remark that the algorithm employs operator $\mathcal{F}_{\mathcal{I}}$ that systematically filters out
the action values of ignorable tokens. More formally, 
\begin{equation*}
    \!\!\mathcal{F}_{\mathcal{I}}[Q (z_{0:t-1}, \cdot)](z)\!=\! 
    \begin{cases}
     Q (z_{0:t-1}, z)  & \!\!\text{if }z \notin \mathcal{I}_{z_{0:t-1}} \\
     -\infty & \!\!\text{if }z \in \mathcal{I}_{z_{0:t-1}}
     \end{cases}\!.
\end{equation*}
This filter operation is used for computing the policy as well as 
for the training target of $Q$-network.

\begin{algorithm}[t]
\caption{PIN (Prompts made INterpretable)} 
\begin{spacing}{0.8}
\begin{small}
    \textbf{Require} : Replay buffer $\mathcal{D}$, parameters $\theta$ for $\psi$ network, parameters $\theta'$ for target $\psi'$ network, and target network update rate $\rho$.\\
    Set parameters of target $\psi'$ network $\theta'$ equal to $\theta$: $\theta' \leftarrow \theta$\\
    \textbf{for} {$iter=1, 2, ... $ } \textbf{do}\\
    \hspace*{4.5mm}\textbf{for} {$t=0, ..., L-1$} \textbf{do}\\ 
    \hspace*{9mm} Sample $z_t \sim \pi_\theta(\cdot | z_{0:t-1})$ in Eq.~\eqref{eq:sp_opt_policy} with\\
    \hspace*{9mm} $\mathcal{F}_{\mathcal{I}}[Q_\theta(z_{0:t-1}, \cdot)]$.\\
    \hspace*{4.5mm}\textbf{end for}\\
    \hspace*{4.5mm}Observe $R(\boldsymbol{y}(\boldsymbol{z}, \boldsymbol{x}))$ from task model.\\
    \hspace*{4.5mm}Add $\{z_0, ... z_{L-1}, R(\boldsymbol{y}(\boldsymbol{z}, \boldsymbol{x})) \}$ to $\mathcal{D}$.\\
    \hspace*{4.5mm}Sample sequence $\{z_0,... z_{L-1}, R(\boldsymbol{y}(\boldsymbol{z}, \boldsymbol{x})) \}$ from $\mathcal{D}$\\
    \hspace*{4.5mm}\textbf{for} {$t=0, 1, ...., L-1$} \textbf{do}\\
    \hspace*{9mm}Estimate the target value $\hat{Q}$ in Eq.~\eqref{eq:sp_target} with\\
    \hspace*{9mm}$\mathcal{F}_{\mathcal{I}}[Q_{\theta'}(z_{0:t}, \cdot)]$.\\
    \hspace*{9mm}Update $\theta$ by minimizing $(Q_\theta(z_{0:t-1}, z_t) - \hat{Q})^2$.\\
    \hspace*{4.5mm}\textbf{end for}\\
    \hspace*{4.5mm}Update target network with $\theta' \leftarrow \rho\theta' + (1 - \rho)\theta$\\
    \textbf{end for}
    \vspace{1.0mm}
\end{small}
\end{spacing}
\label{alg:alg}
\end{algorithm}

\section{Experiments}
\label{sec:exp}

\begin{table*}[t]
    \centering
    \resizebox{\linewidth}{!}{
    \addtolength{\tabcolsep}{3pt}
    \begin{tabular}{lllllllll}
    \toprule
     & \multicolumn{7}{c}{Few-Shot Text Classification Dataset} & \\
     \cmidrule(l{2pt}r{2pt}){2-8}
     \multirow{-2.5}{*}{Method} & SST-2 & Yelp P. & MR & CR & AG's News & Yahoo & Subj & \multirow{-2.5}{*}{Avg.} \\
    \midrule
    Manual Prompt & 82.8 & 83.0 & 80.9 & 79.6 & 76.9 & 18.1 & 51.5 & 67.5 \\
    Instructions & 89.0 & 84.4 & 85.2 & 80.8 & 54.8 & 21.4 & 50.4 & 66.6 \\
    In-Context Demonstration & 85.9 & 89.6 & 80.6 & 85.5 & 74.9 & 36.7 & 73.0 & 77.3  \\
    Soft Prompt Tuning & 73.8 & 88.6 & 74.1 & 75.9 & \textbf{82.6} & \textbf{59.7} & 73.0 & 75.4 \\
    GrIPS & 87.1 & 88.2 & 86.1 & 80.0 & 65.4 & 22.5 & 74.8 & 72.0 \\
    AutoPrompt & 75.0 & 79.8 & 62.0 & 57.5 & 65.7 & 35.5  & 78.9 & 64.9 \\
    $\text{PEZ}^{\dagger}$ & 70.0\,{\small(1.2)} & 85.9\,{\small(0.5)} & 67.9\,{\small(0.4)} & 70.1\,{\small(0.8)} & 43.7\,{\small(0.4)} & 27.0\,{\small(0.7)} & 53.1\,{\small(0.8)} & 59.7 \\
    $\text{RLPrompt}^{\dagger}$ & \underline{91.5}\,{\small(0.2)} & \underline{94.7}\,{\small(0.1)} & \underline{87.1}\,{\small(0.2)} & \textbf{89.5}\,{\small(0.1)} & 77.5\,{\small(0.8)} & 48.6\,{\small(0.1)}  & \underline{82.2}\,{\small(0.4)} & \underline{81.6} \\
    \midrule
    PIN (ours) & \textbf{92.0}\,{\small(0.1)} & \textbf{95.0}\,{\small(0.4)} & \textbf{87.2}\,{\small(0.3)} & \underline{88.3}\,{\small(0.2)} & \textbf{82.6\,{\small(0.1)}} & \underline{49.5}\,{\small(0.3)} & \textbf{85.1\,{\small(0.4)}} & \textbf{82.8} \\
    \bottomrule
    \end{tabular}
}
\caption{Comparison between PIN and baselines on various few-shot text classification datasets. $\dagger$ denotes our reproduced results, and we refer to the results from \citet{rlprompt} for other methods. \textbf{Bold} and \underline{underline} indicate the best and the second best accuracy for each dataset, respectively.
    The accuracy on the test dataset is reported as the average score over 5 few-shot train datasets with the standard error.
    }
\vskip -0.05in
\label{tab:fsc}
\end{table*}

\subsection{Few-Shot Text Classification}
The goal of these tasks is to find the optimal prompt that assigns input text $\boldsymbol{x}$ 
to the class label, given a few examples in the context. 
We employ the same experimental setting in~\citet{schick-schutze-2021-just,rlprompt},
which addresses the classification tasks 
via prompting the task LM and mapping the predicted token to the class label.


%

\paragraph{Experiment Setup}
We employ RoBERTa-large~\cite{roberta} as the task LM 
and OPT-125M as our policy LM.
We experiment with
an extensive range of few-shot text classification task to 
assess the effectiveness of our approach. 
The datasets encompass various domains, including sentiment classification, such as SST-2~\cite{sst2}, Yelp Polarity~\cite{yelp-agnews-yahoo}, MR~\cite{mr}, CR~\cite{cr},
and subjectivity classification like Subj~\cite{subj}.
Furthermore, we also extend to topic classification, such as AG's News and Yahoo~\cite{yelp-agnews-yahoo}. 
%

\paragraph{Baselines}
We compare our algorithm against various baselines, 
including heuristic methods such as:
(1) Manual Prompt, 
(2) Instructions, 
and (3) In-Context Demonstration. 
Furthermore, we consider soft\,/\,hard prompt tuning algorithms, comprising  
(4) Soft Prompt Tuning~\cite{li-liang-2021-prefix} that performs gradient descent directly on continuous embedding vectors,
(5) GrIPS~\cite{prasad-etal-2023-grips} that is a gradient-free and edit-based search method,
(6) AutoPrompt~\cite{shin-etal-2020-autoprompt} that modifies discrete prompts using the gradient information from LM, 
(7) PEZ~\cite{pez} that involves gradient descent on continuous embedding vectors, then projection onto human-readable tokens, and 
(8) RLPrompt~\cite{rlprompt}, the on-policy soft Q-learning, that obviates the need for gradient signal of task LM in the optimization process.

\paragraph{Results}
The results are summarized in Table~\ref{tab:fsc}.
To ensure a fair comparison with RLPrompt,
we use the same reward function and policy LM (i.e., OPT-125M).
As shown in Table~\ref{tab:fsc}, 
our approach demonstrates competitive or superior performance 
compared to RLPrompt across all datasets.
A notable aspect of our PIN method is its efficiency: 
it achieves these strong results 
while necessitating fewer trials than required by RLPrompt.
This is demonstrated in the learning curves, depicted in Figure~\ref{fig:fsc_training_curve} in Appendix.
Furthermore, when compared to various soft\,/\,hard prompt algorithms, 
PIN shows robust performance across the datasets.
This is especially important given that our algorithm relies on a weaker feedback and reward, as opposed to 
the direct back-propagated gradients used in Soft Prompt Tuning, AutoPrompt, and PEZ.
However, PIN underperforms relative to Soft Prompt Tuning on Yahoo dataset.  
The number of classes on Yahoo dataset is 10, which is relatively larger compared to other datasets. Since the reward feedback does not directly reveal the ground-truth label, it poses a harder challenge than using the back-propagated gradient feedback. This is also the reason why RLPrompt did not perform well.

The examples of learned prompts from baselines and PIN are in Table~\ref{tab:fsc_prompts} in Appendix.
Our prompts not only exhibit effectiveness in the task performance but also provide the benefit of enhanced interpretability.

\begin{figure*}[t!]
  \centering
  \includegraphics[width=0.95\linewidth]{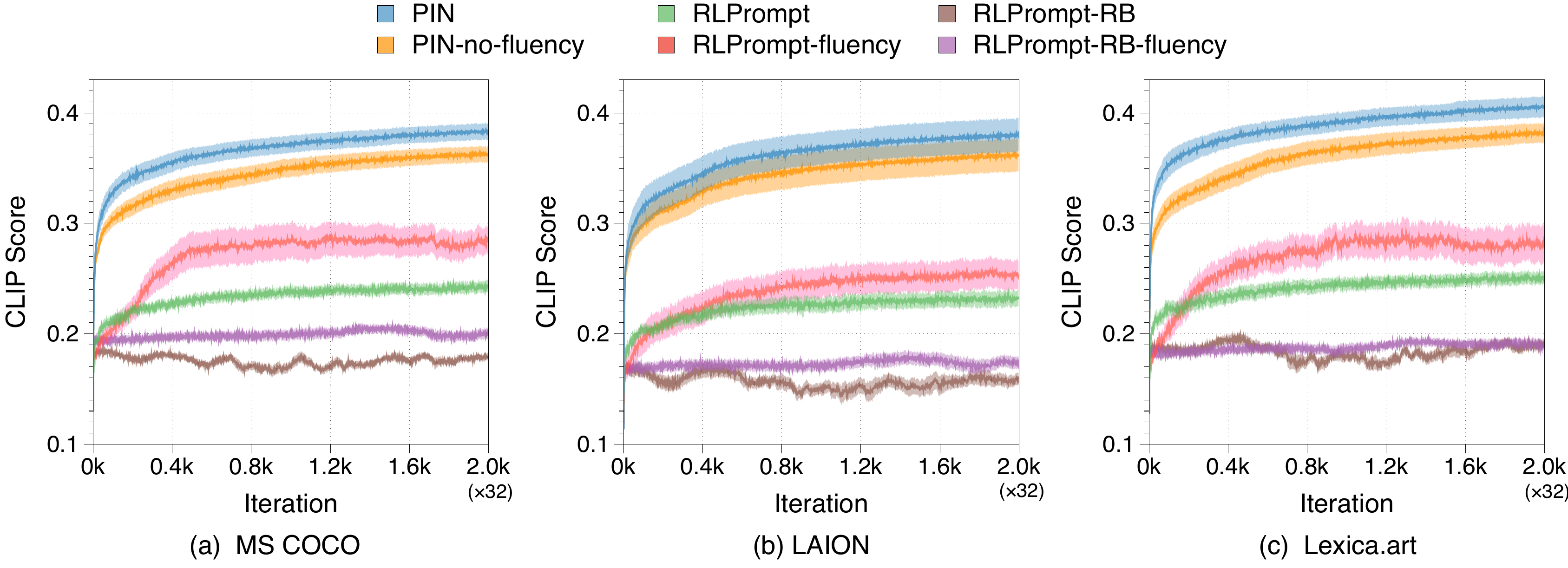}
  \vskip -.1in
  \caption{Training curves on textual inversion from images task. 
  For each dataset, 
  training was conducted on 10 target images across 3 random seeds. 
  The solid curves show the average CLIP Score across target images, with the shaded areas representing the standard error.
  }
  \label{fig:clip_main}
\end{figure*}

\subsection{Unsupervised Text Style Transfer}
The text style transfer task~\cite{DBLP:journals/coling/JinJHVM22} 
aims to rephrase an input text $\boldsymbol{x}$ to match a desired style.
For example, in the sentiment transfer task,
the goal is to alter a negative sentence  
``The movie was disappointing'' into its positive counterpart, 
e.g., ``The movie was awesome''.
We focus on unsupervised text style transfer task, 
where there are no input-output pair examples for training. 

\paragraph{Experiment Setup}
We employ OPT-125M as our policy LM and 
OPT-125M\,/\,350M\,/ 1.3B~\cite{opt} as the task LM.
We conduct the experiments on Yelp~\cite{tst-yelp} that is a widely-used dataset for the sentiment transfer task, focusing on negative-to-positive and positive-to-negative sentiment transfer.
For the rest of the settings, we adhere to the experiment setup outlined in~\citet{rlprompt}.

\paragraph{Baselines}
The output is evaluated by a combined metric that measures content preservation as well as 
style alignment, which is given as the scalar reward feedback. Assuming that the metric 
is not differentiable, prompt tuning methods that directly rely on gradients
are not applicable. 
Therefore, we only compare our method against RLPrompt in this experiment.
Further details regarding our experimental setup and the reward function used in this task can be found in Appendix~\ref{app:tst}.


\paragraph{Results}
The results are summarized in Table~\ref{tab:tst},
where evaluations are conducted on three specific aspects, including content, style, and fluency of output text and two comprehensive metrics like J and GM on test dataset.
For details on the evaluation metrics, refer to Appendix~\ref{app:tst_baselines} and \citet{rlprompt}.
Our experiments, conducted with various sizes of task models under the same policy LM, consistently demonstrate the superiority of our algorithm, as evidenced by the overall metrics, J and GM.
Furthermore, because the reward function primarily focuses on the content and style, our PIN algorithm has notably enhanced performance in relation to these metrics.

However, 
the fluency of the generated text drops in PIN. 
This occurs because the reward function does not account for the generated text fluency, instead it is defined as the sum of content preservation and the target style intensity of the generated text. Thus, there is no guarantee that the fluency will be improved by optimizing the reward, and it may even be possible to achieve higher rewards at the expense of fluency score. We also remark that there is a trade-off, making the generated text aligned closer with the desired style leads to diminished fluency in the generated texts.

\begin{table}[h!]
  \centering
  \resizebox{\columnwidth}{!}{
    \renewcommand{\arraystretch}{1.1}
    \addtolength{\tabcolsep}{-4pt}
  \begin{tabular}{l|lccccc}
    \toprule
    \#\,Param & Method & Content & Style & Fluency & J & GM \\
    \midrule
    \multirow{2}{*}{125M} & PIN & \textbf{55.6}\,\small{(3.2)} & \textbf{95.4}\,\small{(0.3)} & \textbf{89.4}\,\small{(0.9)} & \textbf{46.2}\,\small{(2.3)} & \textbf{77.6}\,\small{(1.1)} \\
    & RLPrompt & 53.5\,\small{(3.4)} & 93.7\,\small{(0.4)} & 85.2\,\small{(2.5)} & 41.4\,\small{(2.1)} & 74.8\,\small{(1.1)} \\    
    \midrule
    \multirow{2}{*}{350M} & PIN & \textbf{58.9}\,\small{(2.7)} & \textbf{94.2}\,\small{(0.8)} & 83.4\,\small{(5.7)} & \textbf{46.8}\,\small{(4.3)} & \textbf{76.0}\,\small{(2.7)}  \\
    & RLPrompt & 52.1\,\small{(1.3)} & 93.9\,\small{(0.3)} & \textbf{85.3}\,\small{(1.7)} & 41.0\,\small{(0.7)} & 74.6\,\small{(0.5)} \\
    \midrule 
    \multirow{2}{*}{1.3B} & PIN & \textbf{69.1}\,\small{(2.7)} & \textbf{95.7}\,\small{(0.5)} & 84.9\,\small{(1.6)} & \textbf{56.0}\,\small{(3.3)} & \textbf{82.4}\,\small{(1.6)} \\
    & RLPrompt & 67.6\,\small{(2.4)} & 91.6\,\small{(1.1)} & \textbf{89.2}\,\small{(0.9)} & 54.1\,\small{(2.4)} & 81.9\,\small{(1.0)} \\
    \bottomrule
  \end{tabular}
}
\caption{Comparison between PIN and RLPrompt on different sizes of OPT model. The reported scores are the average values with the standard error of 4 different runs for each transfer task. We assess the output text from the task LM using various metrics.
}
  \label{tab:tst}
  \vskip -.07in
\end{table}

\begin{figure*}[ht]
  \centering
  \includegraphics[width=\linewidth]{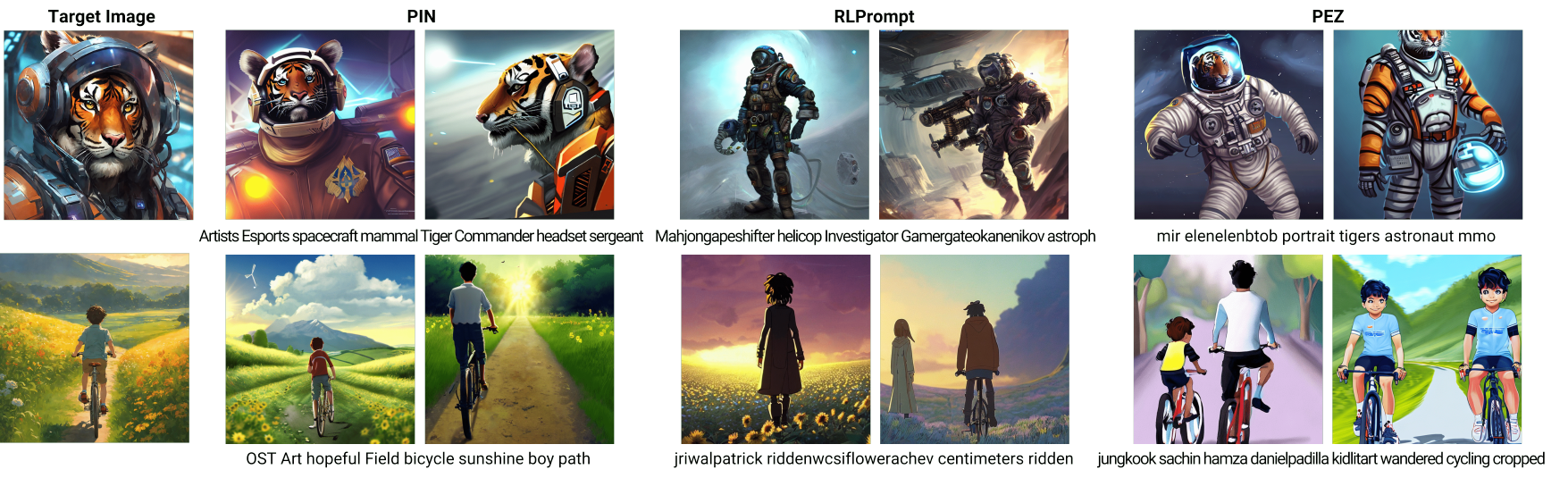}
  \caption{Generated images from Stable Diffusion-v2~\cite{latent-diff}, using the learned hard prompts (bottom) for the target image (left). We also showcase more qualitative examples in Appendix~\ref{app:qau}.}
  \vskip -.02in
  \label{fig:clip_tiger}
\end{figure*}

\subsection{Textual Inversion From Images}
\label{sec:pii}
Hard prompts are also useful in the vision-language domains~\cite{pez}.
A salient task in these domain is textual inversion, which entails identifying the caption that describe target images using VLMs, such as CLIP~\cite{clip}.
The caption can then serve as a prompt for generating similar images via text-guided
diffusion models.
To learn the prompt,  
we make use of the CLIP Score~\cite{hessel2021clipscore} as the reward that measures the similarity between
the target image and the prompt.

\paragraph{Experiment Setup}
We set OpenCLIP-ViT/H~\cite{DBLP:conf/cvpr/ChertiBWWIGSSJ23} as the task model and OPT-350M as our policy LM.
We use a range of image datasets for textual inversion,
including MS COCO~\cite{coco}, LAION~\cite{laion-400m}, and Lexica.art~\cite{lexica}.
For each dataset, we randomly select
10 target images to learn the prompt, and calculate the average CLIP Score of the target image and the generated prompt over 3 seeds.
For the rest of the settings, we adhere to the experimental setup outlined in~\citet{pez}.

\paragraph{Baselines}
For the evaluation of training efficiency, we compare the training curves against a range of
RL-based algorithms. 
These baselines include:
(1) RLPrompt, 
(2) RLPrompt-fluency, a variant of RLPrompt with a filtering method akin to our own,
(3) RLPrompt-RB, which incorporates the replay buffer,
(4) RLPrompt-RB-fluency, which incorporates both filtering and the replay buffer,
(5) PIN-no-fluency, which trains Q-network with the sparse Tsallis entropy regularization but without filtering.
Further details about these baselines can be found in Appendix~\ref{app:ti_baselines}.

\paragraph{Results}
The training curves, as depicted in Figure~\ref{fig:clip_main}, 
illustrate that PIN is most effective in discovering high-quality prompts with fewer interactions, particularly
compared to RLPrompt.
This efficiency is primarily due to addressing 
the overdetermined issue discussed in Section~\ref{sec:issue}.
Furthermore, when compared with RLPrompt-fluency, which employs the same filtering operator
to eliminate unlikely tokens, 
PIN clearly demonstrates better performance. This can be explained as follows:
To ensure that RLPrompt-fluency works properly, we had to filter out tokens aggressively
due to the dense nature of the softmax distribution. 
This resulted in excluding tokens that are potentially important for task performance.
In contrast, 
the sparse Tsallis entropy regularization employed by PIN
enabled to handle a larger token search space. 

On the other hand, we note that RLPrompt-RB and RLPrompt-RB-fluency struggle to learn the prompt. 
At first glance, it is a surprising contradiction to the common practice in
deep RL that the replay buffer generally improves performance.
However, this is a very natural result: replay buffers help when training non-linear function approximators such as 
deep neural networks that can potentially overfit. Since the $Q$-network is designed to resolve an exceedingly overdetermined linear system, 
the replay buffer offers minimal benefit. Instead, the weights in the 
least squares formulation over tokens in Eq.~\eqref{eqn:over} correspond to the mixed distribution of previous policies during training, 
including poor ones. Thus, the problem with the overdetermined linear system for training the $Q$-network is
compounded by the fact that some effort will be put on reducing the approximation error for tokens that are now found to be 
unimportant, in sacrifice of accurate estimation for important tokens. We suspect that this is why the original implementation of 
RLPrompt, although based on SQL that employs replay buffer, does not include the replay buffer.
RLPrompt-RB-fluency falls short in discovering effective prompts due to their limited search space as mentioned earlier. 
In contrast, PIN and PIN-no-fluency do not suffer from the aforementioned problem related to replay buffers.
We provide some qualitive examples of the generated images and learn prompts in Figure~\ref{fig:clip_tiger}.

\begin{table}[t!]
  \centering
  \resizebox{\columnwidth}{!}{
    \renewcommand{\arraystretch}{1.03}
    \addtolength{\tabcolsep}{1pt}
  \begin{tabular}{l|cc|cc}
    \toprule
     & \multicolumn{2}{c|}{Relevance} & \multicolumn{2}{c}{Interpretability} \\
     \cmidrule(l{2pt}r{2pt}){2-3} \cmidrule(l{2pt}r{2pt}){4-5}
     \multirow{-2.5}{*}{Dataset} & PIN & PEZ & PIN & PEZ \\
    \midrule
    MS COCO & \textbf{3.29}\,\small{(0.17)} & 2.53\,\small{(0.24)} & \textbf{2.29}\,\small{(0.16)} & 1.70\,\small{(0.18)} \\    
    \midrule
    LAION & \textbf{4.46}\,\small{(0.18)} & 3.33\,\small{(0.23)} & \textbf{3.60}\,\small{(0.20)} & 2.10\,\small{(0.21)} \\    
    \midrule
    Lexica.art & \textbf{4.30}\,\small{(0.12)} & 2.90\,\small{(0.34)} & \textbf{3.40}\,\small{(0.18)} & 2.10\,\small{(0.27)} \\   
    \bottomrule
  \end{tabular}
}
\caption{Human-like evaluation on PIN and PEZ by using the GPT-4V API. 
The reported numbers are average over 10 target images with the standard error for each dataset. We omitted the RLPrompt results because all the scores were 1.
}
  \label{tab:ai_eval}
  \vskip -.1in
\end{table}

\paragraph{Human-like Evaluation}
For evaluating the interpretability of the learned hard prompts 
and their relevance to the target images, 
we adopt Human-like ChatGPT evaluation method inspired by \citet{chatgpt-eval}. 
We request GPT-4V to assign a score ranging from 1\,(worst) to 5\,(best), assessing the relevance between the target image and the hard prompt, as well as the prompt's interpretability for human understanding. We provide the template for this in Figure~\ref{fig:template} in Appendix.
To provide a comparative baseline, 
we conduct the same evaluation with prompts learned by PEZ.
The results, presented in Table~\ref{tab:ai_eval},
suggest that the prompts learned by PIN are more interpretable to humans and more accurately capture the content of the target images.


\subsection{Analysis on Hyperparameters}

\begin{figure}[!h]
\centering
\subfigure[Analysis on prompt length]{\label{fig:abl_prompt_len}\includegraphics[width=0.51\columnwidth]{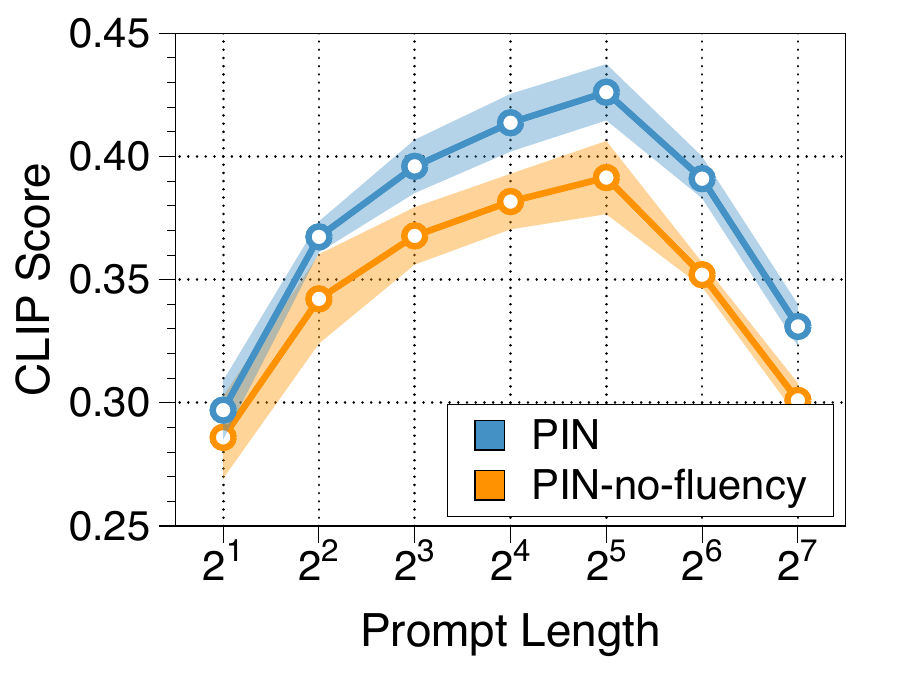}}
\subfigure[Analysis on top-k]{\label{fig:abl_top-k}\includegraphics[width=0.47\columnwidth]{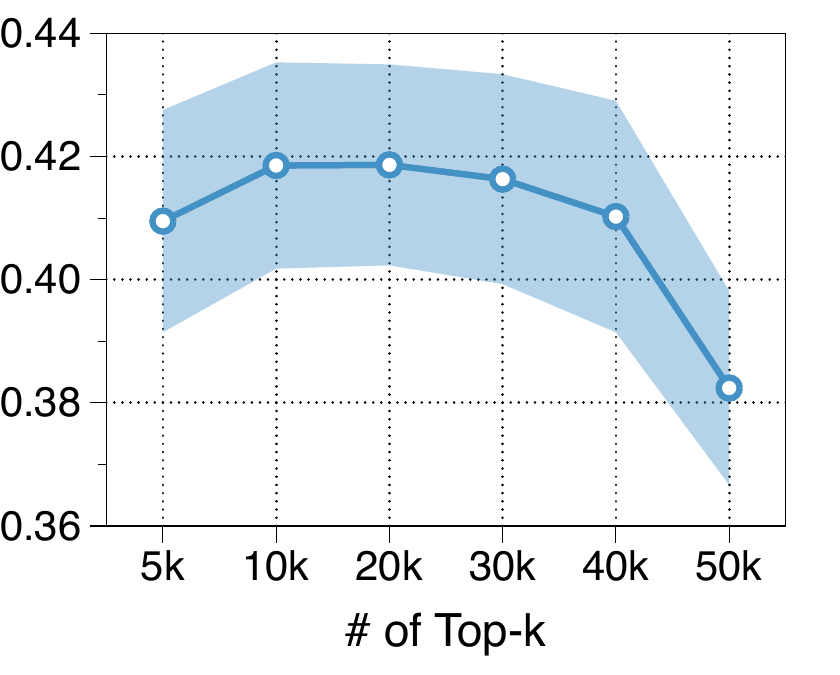}}
    \caption{(a) Comparison with PIN-no-fluency at varying prompt length, and (b) Analysis on the effect of $k$.}
    \label{fig:ablation_study}
\end{figure}

\paragraph{Length $L$ of Prompts}
To further evaluate the effectiveness of our algorithm relative to PIN-no-fluency,
we conducted experiments with varying lengths of prompts.
The results, as illustrated in Figure~\ref{fig:abl_prompt_len}, reveal that the performance gap between our method and PIN-no-fluency widens with increasing prompt length up to $L=2^5$.
There are two primary reasons for this observed trend.
Firstly, our algorithm reduces the search action space to a set of familiar tokens.
As prompt length increases, the search space expands exponentially; 
thus, our approach to narrowing the action space results in enhanced performance compared to PIN-no-fluency.
Secondly, PIN-no-fluency attempts to estimate the ground- truth Q-value across all tokens at every state.
With increasing prompt length, the state space also expands, which can lead to inaccuracies in Q-value estimation at each state.
Our method, by contrast, mitigates this challenge, leading to more reliable and effective prompt optimization, particularly in scenarios with longer prompts.

For prompts of length $L=2^5$, PIN achieves the highest Clip Score. 
However, for $L=2^6, 2^7$, the score drops. 
We remark that longer prompts pose the challenge of combinatorial search space of hard prompts for RL-based methods, and we believe that we need more information-rich feedback other than just reward signals for making the methods more effective. This would be nontrivial for our setting where the LM is assumed to be a black-box model, but it is a promising direction for future work. 

\paragraph{$k$ For Determining $\mathcal{I}$}
An important determinant of the performance of our algorithm is the number of tokens ignored at each state.
We conduct an analysis focusing on the hyperparameter $k$, which represents the number of tokens considered for Q-values estimation.
The findings are presented in Figure~\ref{fig:abl_top-k},
where the last datapoint, indicating $50k$ is the same with PIN-no-fluency.
A smaller action space ($|\mathcal{I}| \approx \mathcal{V}$), 
can exclude tokens that are crucial for discovering optimal prompts.
Such an exclusion risks limiting our algorithm's ability to identify the most effective prompts as potentially valuable tokens could be filtered out prematurely.
Conversely, an excessively large action space ($|\mathcal{I}| \ll  \mathcal{V}$) from fewer ignored tokens, diminished the impact of our filtering technique.
In such cases, the benefit of reducing search space is lost, 
as the algorithm still needs to evaluate a vast number of tokens.
Our empirical investigations across various tasks indicate 
that maintaining the number of tokens that are not to be ignored within the range of 10000 to 20000 yields the most favorable results, 
particularly when the coefficient $\alpha=1$. 

\section{Limitations}
PIN can discover prompts that are more interpretable compared to baselines.
However, we acknowledge a couple of inherent limitations. 
Firstly, the algorithm exhibits a relatively higher time consumption for training when compared to gradient-based methods. Secondly, despite its advanced capabilities in discovering interpretable prompts, 
our algorithm does not guarantee the consistent discovery of grammatically perfect sentences.
We would leave discovering grammatically perfect prompts as future work.

\section{Conclusion}
\label{sec:conclusion}
In this paper, 
we firstly discuss the overdetermined issue encounterd in the efficiently parameterized network.
To address this issue,
we propose PIN algorithm, 
which uses sparse Tsallis entropy regularization to systematically exclude ignorable tokens 
from constraints.
Prompts learned by PIN 
exhibit better performance in various tasks. 
Future work could explore alternative strategies for identifying ignorable tokens to improve interpretability further, like leveraging task-specific domain knowledge.

\section*{Acknowledgments}
This work was supported by Institute for Information \& communications Technology Promotion (IITP) grant funded by MSIT (No.RS-2019-II190075 Artificial Intelligence Graduate School Program(KAIST); 
No.2020-0-00940 Foundations of Safe Reinforcement Learning and Its Applications to Natural Language Processing; 
No.RS-2024-00343989 Enhancing the Ethics of Data Characteristics and Generation AI Models for Social and Ethical Learning).

\bibliography{ref, custom}
\clearpage

\appendix
\clearpage

\section{Hyperparameter Settings}
We employ 2 MLP layers with 2048 hidden states for the implementation of $\psi$.
During the learning of our PIN method,
we sample a batch of 256 sequences from the replay buffer to update the parameters of Q-network.
We use an Adam optimizer with learning rate 5e-5.
Note that we maintain consistency with PIN and other RL algorithms in all shared hyperparameters, such as the learning rate, except for the reward scale, which varies across tasks.
Our experiments mainly follow the setting in~\citet{rlprompt}.

\section{Few-Shot Text Classification}
\label{app:fsc}
The classification process begin with integrating the input text $\boldsymbol{x}$ that needs to be classified and the prompt $\boldsymbol{z}$ into a templated format: `[$\boldsymbol{x}$] [$\boldsymbol{z}$] [MASK]'. 
Subsequently, 
the classification decision relies on selecting the predefined tokens, each representing a specific class, 
that has the highest probability of filling the [MASK] position.
\subsection{Experiment Setup}
We use OPT-125m ($\text{dim}(\mathcal{E})=768$, $|\mathcal{V}|=50272$)
as our backbone of policy-LM, and prompt length $L$ is 5 over all experiments.
During training, the prompt is learned from a training dataset containing a few pairs of input text $\boldsymbol{x}$ and their corresponding labels $c$.
Additionally, a validation dataset is used to assess the prompts during training.
The accuracy of the predicted labels is evaluated on a test dataset
using the prompt that demonstrated the best performance on the validation dataset.
For each dataset, 
we sample 5 distinct sets for training and validation, each includes 16 examples per class. 
We run 3 experiments with a different random seed for each set and we report the average accuracy.
For our PIN algorithm, 
the reward scale ($1/\alpha$) is 1 and $k$ is 10000 ($|\mathcal{I}|$ is 40272).
\begin{figure*}[h]
  \centering
  \includegraphics[width=0.85\linewidth]{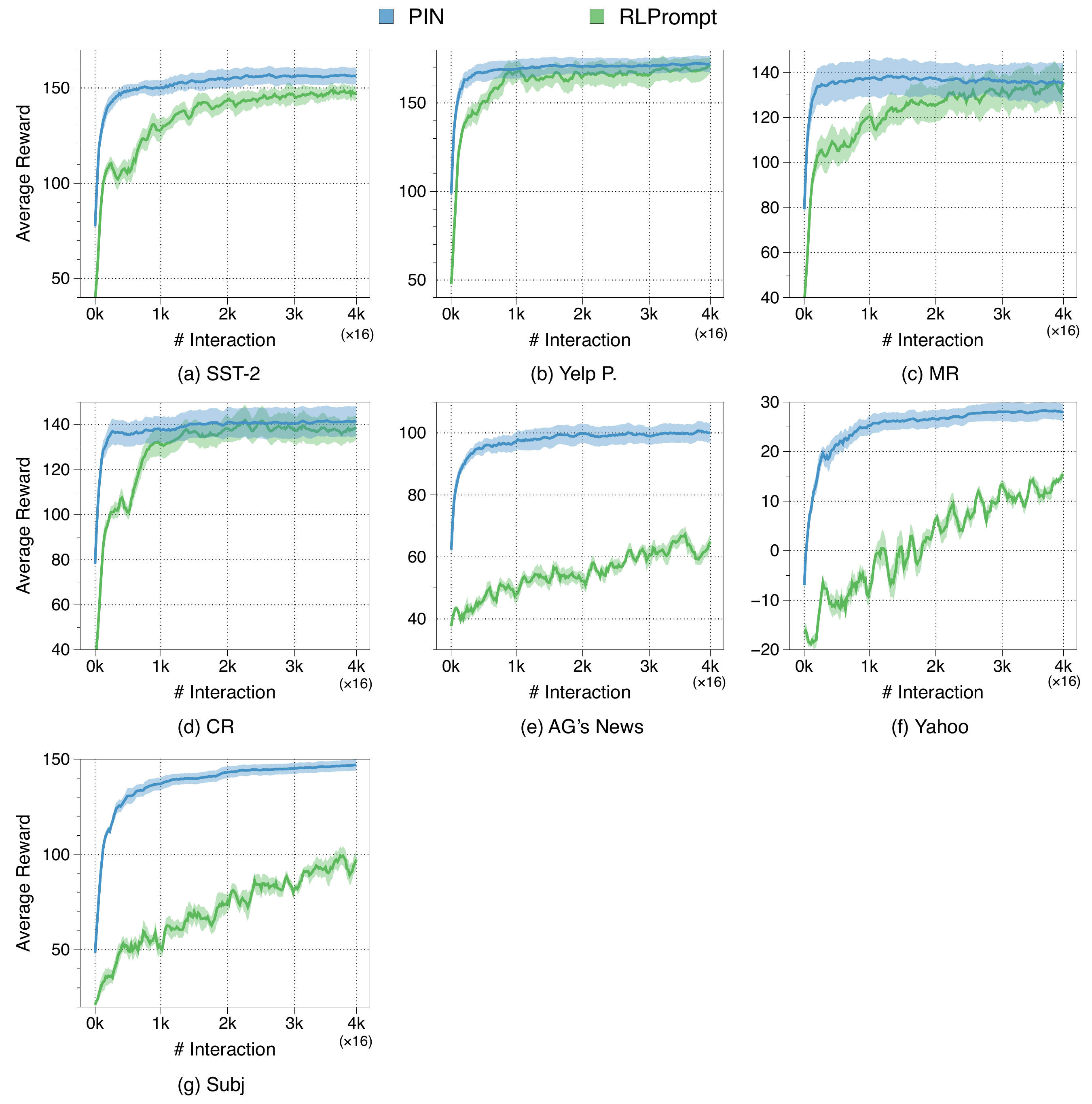}
  \caption{    
    Training curve between PIN (ours) and RLPrompt.
    For each of the 5 few-shot training sets, 3 experiments were conducted.
    The graph depicts the average reward over the validation sets with standard error shading.}
  \label{fig:fsc_training_curve}
\end{figure*}

\subsection{Rewards}
The objective of the text classification task is to accurately assign input text
$\boldsymbol{x}$ to its corresponding ground truth label $c$.
we employ a piecewise reward function designed to incentivize the correct classification of each example. 
We use a piecewise reward function designed to enhance prompts 
to classify each example correctly as used in \citet{rlprompt}.
For a given prompt $\boldsymbol{z}$ and a training example $(\boldsymbol{x}, c)$, 
we compute the reward in a manner akin to hinge loss.
This is achieved by measuring the gap between the probability of the correct label and the highest probability among the other classes.
we denote 
$P_{\text{LM}}(c|\boldsymbol{z}, \boldsymbol{x})$ 
as the probability of label $c$, 
we can write the gap as 
$\text{Gap}_{\boldsymbol{z}}(c) := P_{\text{LM}}(c|\boldsymbol{z}, \boldsymbol{x}) - \max_{c'\neq c}P_{\text{LM}}(c'|\boldsymbol{z}, \boldsymbol{x})$.
This gap is positive when the prediction is correct and negative otherwise. 
We define a binary indicator for correct predictions as $\text{Correct} := \mathbbm{1}[\text{Gap}_{\boldsymbol{z}}(c) > 0]$.
For correct predictions, we amplify the positive reward by a larger factor to emphasize its desirability.
The reward function is thus formulated as follows:
\begin{equation*}
    R(\boldsymbol{x}, c) = 
    \lambda_1^{1-\text{Correct}}
    \lambda_2^{\text{Correct}}
    \text{Gap}_{\boldsymbol{z}}(c)
\end{equation*}
and $\lambda_1=180, \lambda_2=200$.

\subsection{Baselines}
We retrained PEZ~\cite{pez} and RLPrompt~\cite{rlprompt} using the official repositories\footnote{https://github.com/YuxinWenRick/hard-prompts-made-easy}\footnote{https://github.com/mingkaid/rl-prompt}.
For PEZ, we used 5 prompt tokens and conducted training with a batch size 16 with the few-shot train dataset.
For a fair evaluation, we utilized the same policy LM for training RLPrompt as was used in our method.
For RLPrompt, we followed standard setting~\cite{rlprompt} by setting the reward scale $(1/\alpha)$ to 5 and using top-256 sampling from the prompt policy during training.
Note that top-256 sampling is performed based on estimated Q-values over tokens. 
The performance results of the other methods are presented based on the reported in~\citet{rlprompt}.

\subsection{Learning Curve}
Figure~\ref{fig:fsc_training_curve} shows the reward 
for the learned prompt on the validation dataset during training.
It demonstrates that our algorithm outperforms RLPrompt across all datasets and requires fewer interactions with task model.

\section{Unsupervised Text Style Transfer}
\label{app:tst}
\subsection{Experiment Setup}
We use OPT-125m ($\text{dim}(\mathcal{E})=768$, $|\mathcal{V}|=50272$)
as our backbone of policy-LM, and prompt length is 5 over all experiments.
The reward scale ($1/\alpha$) is 2 and $|\mathcal{I}|$ is 40271.
Our experiments follow the suggested setting of text style transfer task in~\cite{rlprompt}.
Dataset Statistics Yelp~\cite{tst-yelp}
contains 266K positive and 177K negative reviews for training, 
38K and 25K for validation, and 76K and 50K for testing, respectively. 
We perform evaluation on a separate dataset consisting of 500 reviews for each sentiment, 
with reference outputs collected by~\citet{li-etal-2018-delete}. 

\subsection{Reward}
Given input text $\boldsymbol{x}$, 
the goal of this task is to generate output $\boldsymbol{y}$ 
that not only preserves the information from $\boldsymbol{x}$ but also aligns 
with the desired style, denoted by $s$.
To quantify the success in achieving these objectives, 
we define the task reward as a simple sum of content preservation and
target style intensity, described formally below:
\begin{equation*}
    R(\boldsymbol{x}, \boldsymbol{y}, s) = \text{Content}(\boldsymbol{x}, \boldsymbol{y}) + \text{Style}(\boldsymbol{y}, s)
\end{equation*}
We implement 
our content preservation reward using its CTC metric~\cite{deng-etal-2021-compression}, which measures the bi-directional information alignment between input
$\boldsymbol{x}$ and output $\boldsymbol{y}$. 
For the style reward, we compute the target style probability under a BERT-base-uncased classifier learned from the training data.

\subsection{Baselines}
\label{app:tst_baselines}
For RLPrompt, we set the reward scale $(1/\alpha)$ to 80 and 
using top-50 sampling based on Q-values from learned Q-network as the suggestion in the origin paper. 
We evaluated performance using five different evaluation metrics.
`Content' is the degree of content preservation
between input text and output text 
based on \citet{deng-etal-2021-compression}. 
`Style' is measured by fine-tuned style classifiers, and `fluency' is assessed by a grammatical acceptability classifier~\cite{krishna-etal-2020-reformulating}.
Additionally, we calculate a joint sentence-level score (J) across test dataset, 
following \citet{krishna-etal-2020-reformulating}, 
and the geometric mean (GM) of the three aspects score.

\begin{figure*}[!t]
    \centering
    \includegraphics[width=0.65\linewidth]{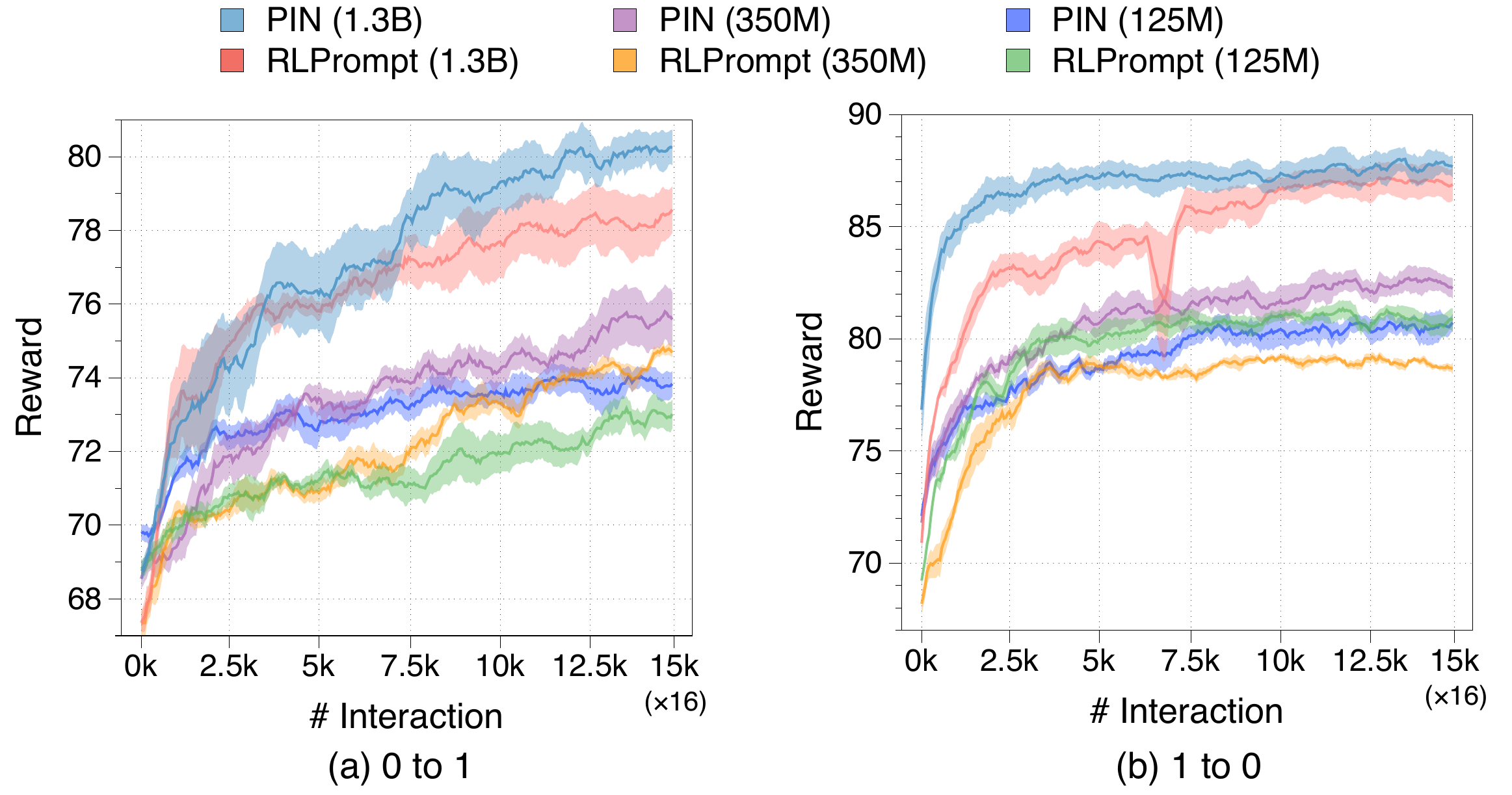}
    \caption{
    Training curve between our PIN method and RLPrompt over various task models (e.g. OPT-125M / 350M / 1.3B).
    For each dataset, specifically for sentiment style transfer including negative-to-positive (0 to 1) and positive-to-negative (1 to 0), 4 experiments were conducted with different seeds.
    The graph depicts the average reward on the validation set with standard error shading.}
    \label{fig:tst}
\end{figure*}

\subsection{Learning Curve}
Figure~\ref{fig:tst} shows the average reward 
for the learned prompt on the validation dataset during training.
It demonstrates that our algorithm outperforms RLPrompt across all dataset and task models, with the exception of 
the positive-to-negative case on OPT-125M task LM.

\begin{figure*}[t!]
\centering
\subfigure[Analysis on reward scale $1/\alpha$ ]{\label{fig:app_ti_a}\includegraphics[width=0.31\linewidth]{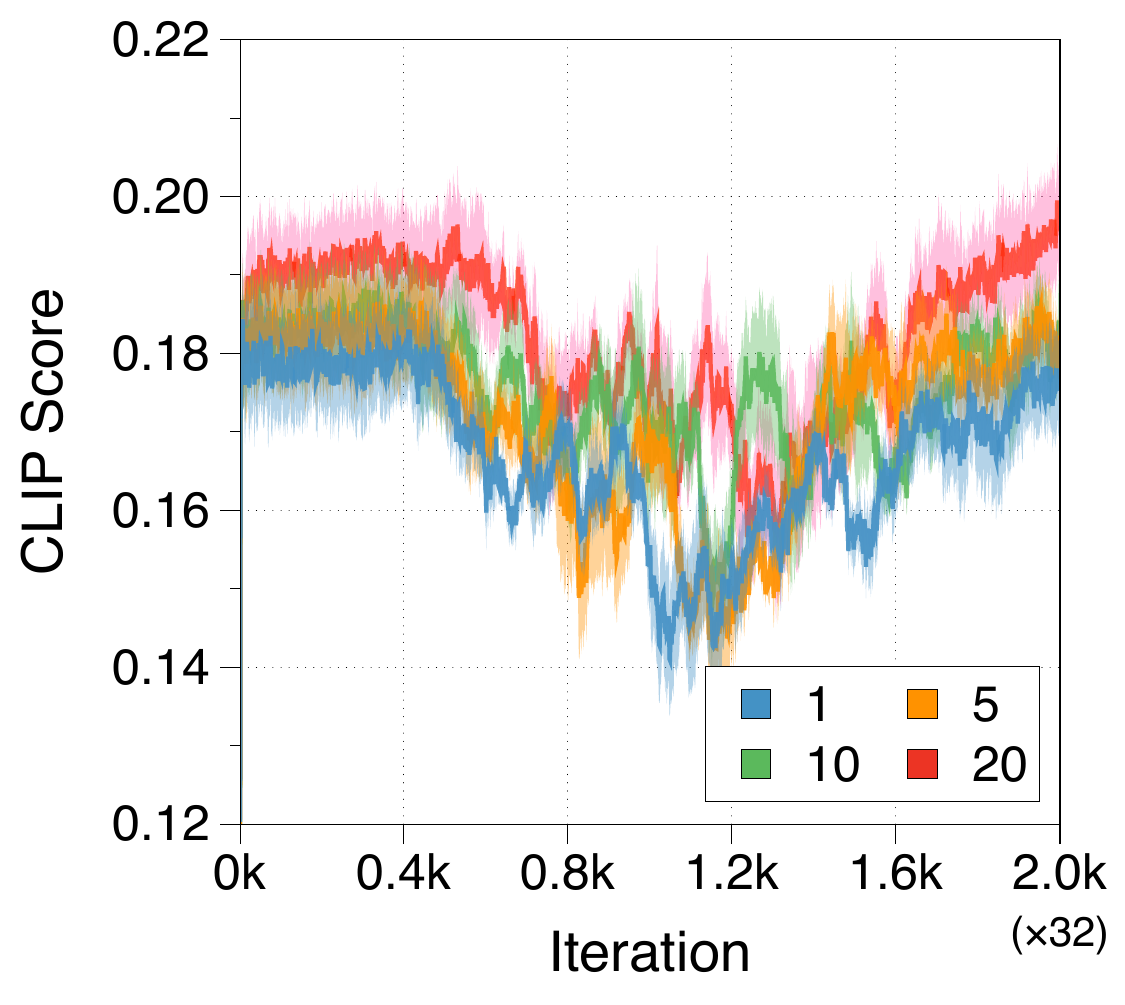}}
\subfigure[Analysis on fluent top-$k$]{\label{fig:app_ti_b}\includegraphics[width=0.31\linewidth]{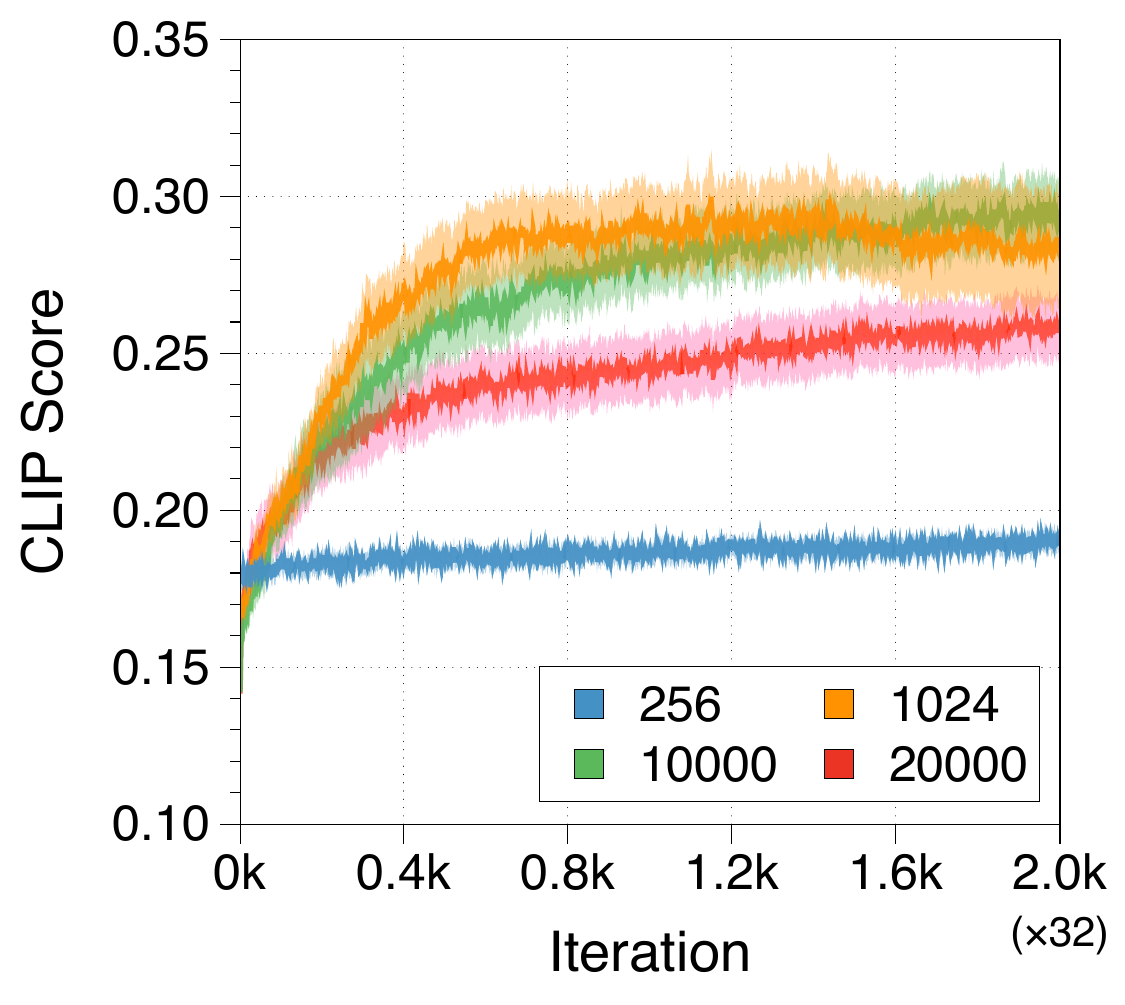}}
\subfigure[Analysis on fluent top-$k$]{\label{fig:app_ti_c}\includegraphics[width=0.31\linewidth]{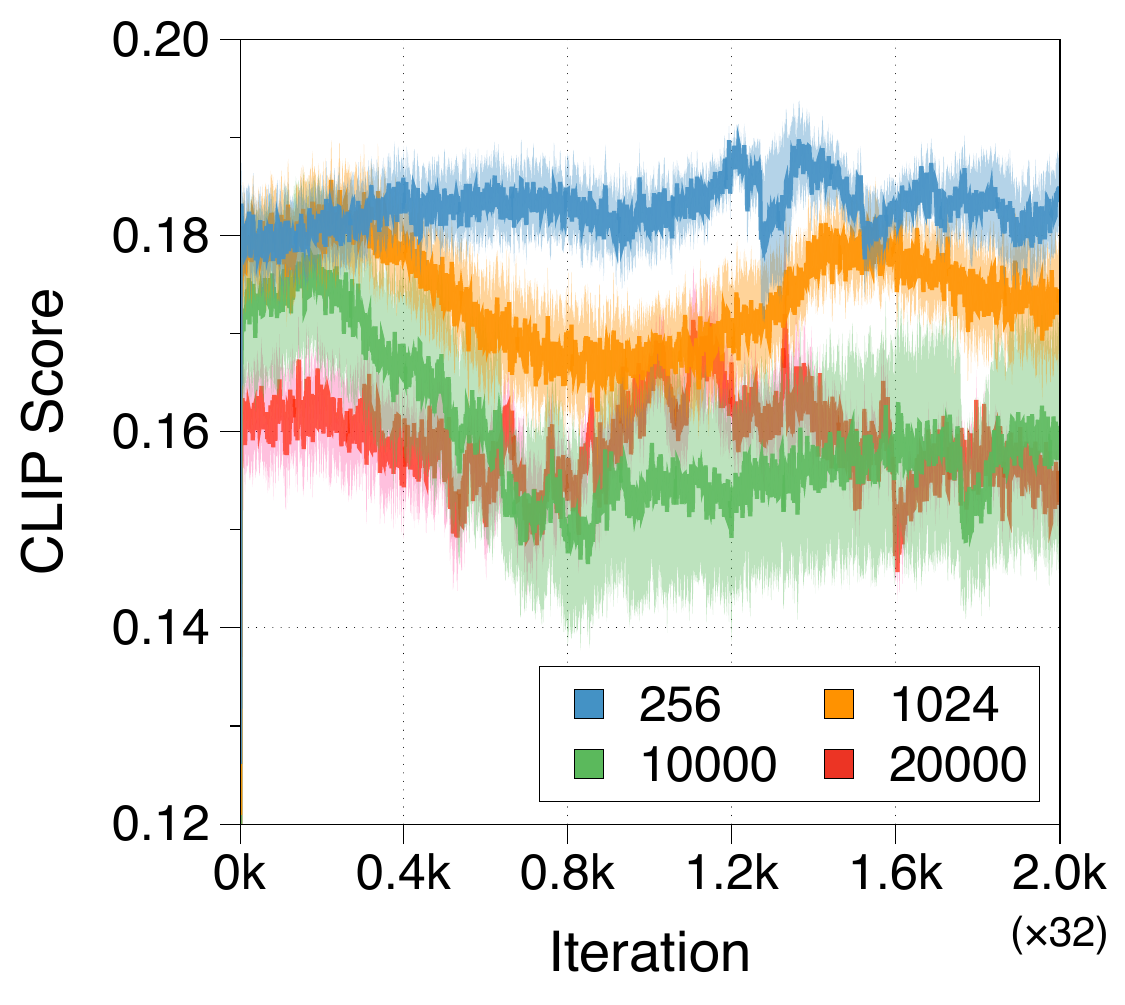}}
    \caption{(a) Analysis on reward scale ($1/\alpha$) in RLPrompt-RB, 
    (b) Analysis on fluent top-$k$ in RLPrompt-fluency and,
    (c) Analysis on fluent top-$k$ in RLPrompt-RB-fluency.}
    \vskip -0.1 in
    \label{fig:app_ti_ablation_study}
\end{figure*}

\section{Textual Inversion from Images}
\label{app:ti}
\subsection{Experiment Setup}
We use OPT-350m ($\text{dim}(\mathcal{E})=1024$, $|\mathcal{V}|=50272$)
as our backbone of policy-LM, and prompt length is 8 over all experiments.
The reward scale ($1/\alpha$) is 1 and $|\mathcal{I}|$ is $30271$ ($k=20000$).
PIN-no-fluency use the same reward scale ($1/\alpha$).
For image generation, 
Stable Diffusionv2~\cite{latent-diff} is utilized.
In the configuration of Stable Diffusion-v2, 
we set the guidance scale to 9 and the number of inference steps to 25.

\subsection{Reward}
We use the Clip Score as a reward.
Formally, given an image encoder function $f: \mathcal{X} \to \mathcal{E'}$ for a target image $\boldsymbol{x} \in \mathcal{X}$ and a text encoder function $g: \mathcal{Z} \to \mathcal{E'}$ for prompt $\boldsymbol{z} \in \mathcal{Z}$,
where $\mathcal{E'}$ denotes the shared embedding space in VLMs,
we define reward function as the cosine similarity between two vectors $f(\boldsymbol{x})$ and $g(\boldsymbol{z})$.

\subsection{Baselines}
\label{app:ti_baselines}
\paragraph{RLPrompt and RLPrompt-fluency}
For RLPrompt and RLprompt-fluency, we set the reward scale ($\alpha$) as 80. 
This variant of RLPrompt incorporates an additional sampling strategy during data collection. Similar to our algorithm, 
RLPrompt-fluency selects from the top-$k$ tokens that exhibit high logits in $W^{\text{LM}}e_t$.
\citet{rlprompt} also investigated this technique in the paper.
In our experiments, 
we investigated various settings for $k$, specifically $k\in \{256, 1024, 10000, 20000\}$.
The performance outcomes for each value $k$ are depicted in Figure~\ref{fig:app_ti_b}.
We observed that a relatively small value of $k$ (e.g. 256) restricts the search space,
thereby hindering the discovery of optimal hard prompts.
Our finding was that the most proper hyperparameter was $k=10000$.
Beyond this point, we show a performance degradation, 
likely attributable to issues arising from Section~\ref{sec:issue}.
It is noteworthy that while our algorithm uses $k=20000$ in this task, 
the use of this value in RLPrompt-fluency did not yield the best results.
This discrepancy suggests that our algorithm is more proficient in estimating Q-values and exploring a larger token space.
We set $k=10000$ over datasets for Figure~\ref{fig:clip_main} in this algorithm.

\paragraph{RLPrompt-RB}
RLPrompt-RB, in our experiments, predominantly exhibited failure in training.
We conducted experiments on various settings for the value of reward-scale (i.e. $1/\alpha$).
The results are depicted in Figure~\ref{fig:app_ti_a}.
We set reward-scale as $20$ over datasets for this algorithm in Figure~\ref{fig:clip_main}.

\paragraph{RLPrompt-RB-fluency}
RLPrompt-RB-fluency 
incorporates a selective sampling strategy during data collection, as described in RLPromopt-fluency. 
Specifically, this variant samples only from the top-$k$ tokens, 
subsequently storing them in the replay buffer $\mathcal{D}$.
In our experiments, 
we explored a range of settings for $k$, specifically $k\in \{256, 1024, 10000, 20000\}$.
The impacts of these different $k$ values on performance are presented in Figure~\ref{fig:app_ti_c}.
Our observations revealed that a smaller value of $k$, such as 256, yields the best results.
Conversely, as the number of $k$ increases, 
we noted a trend towards instability in the training process. 
This phenomenon is discussed in Section~\ref{sec:issue}.
Thus, we set hyperparameter, $k$ as $256$ over datasets for this algorithm in Figure~\ref{fig:clip_main}.
The reward scale ($1/\alpha$) is 20.

\subsection{Human-like Evaluation}
\begin{figure}[t!]
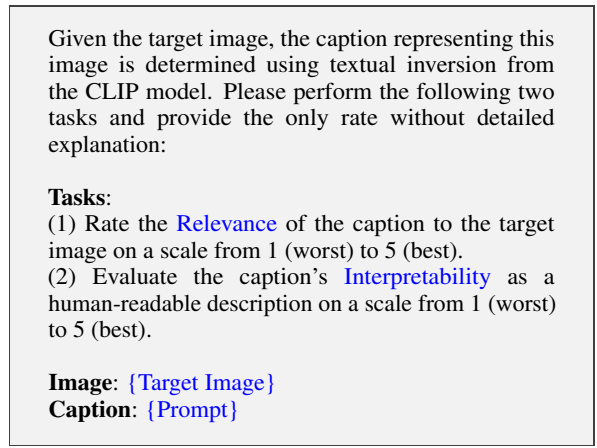

    \centering
    \small
    \begin{tcolorbox}
    [width=\linewidth, sharp corners=all, colback=gray!10, boxrule=0.2mm]
    Given the target image, the caption representing this image is determined using textual inversion from the CLIP model. Please perform the following two tasks and provide the only rate without detailed explanation:\\

    \textbf{Tasks}:\\
    (1) Rate the \textcolor{blue}{Relevance} of the caption to the target image on a scale from 1 (worst) to 5 (best). \\
    (2) Evaluate the caption's \textcolor{blue}{Interpretability} as a human-readable description on a scale from 1 (worst) to 5 (best).\\

    \textbf{Image}: \textcolor{blue}{\{Target Image\}} \\
    \textbf{Caption}: \textcolor{blue}{\{Prompt\}}
    \end{tcolorbox}
    \vspace{-5pt}
    \caption{The template for human-like evaluation to score relevance and interpretability of the hard prompts. We utilized GPT-4V APIs.}
    \label{fig:template}
\end{figure}

We select prompts that achieve the highest Clip Score during the training 
in both PEZ and PIN algorithms.
These selected prompts are then evaluated in GPT-4V using the template provided in Figure~\ref{fig:template}.

\section{Qualitative Examples}
\label{app:qau}
\begin{figure*}[h]
    \centering
    \includegraphics[width=\linewidth]{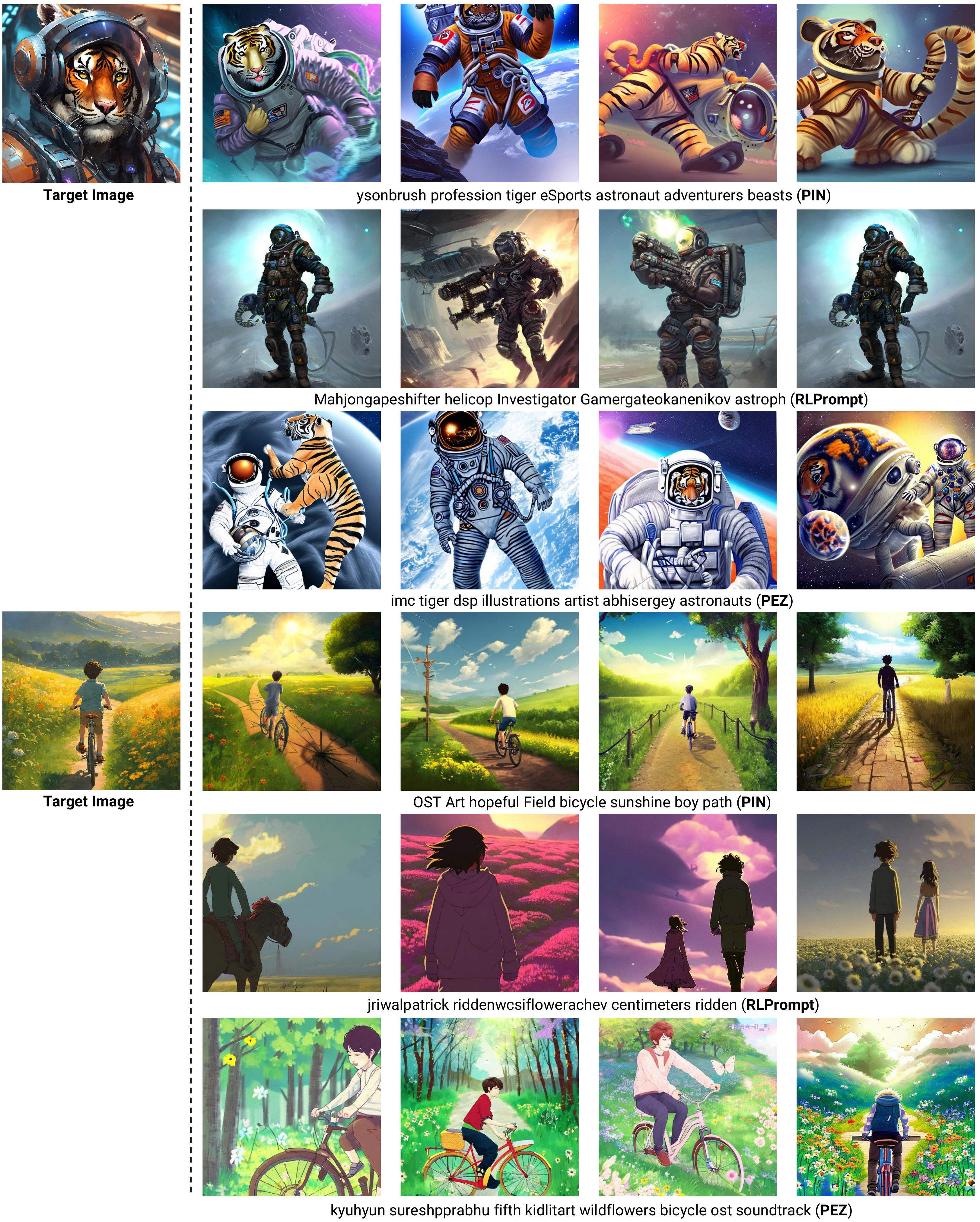}
    \caption{Generated images using learned hard prompts (bottom) through Stable Diffusion-v2~\cite{latent-diff} for a given target image (left).}
    \label{fig:clip_ex_v1}
\end{figure*}

\begin{figure*}[h]
    \centering
    \includegraphics[width=\linewidth]{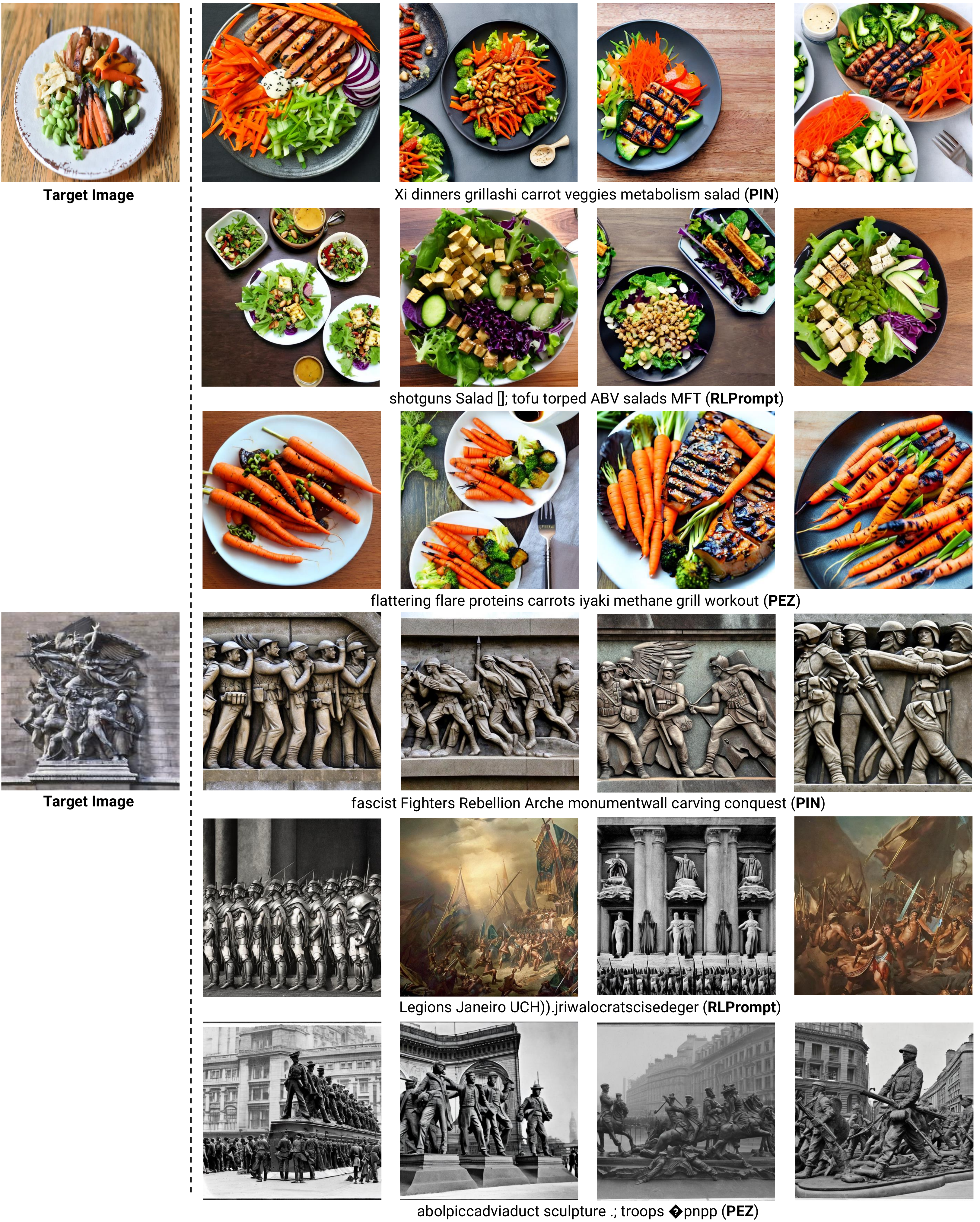}
    \caption{Generated images using learned hard prompts (bottom) through Stable Diffusion-v2~\cite{latent-diff} for a given target image (left).}
    \label{fig:clip_ex_v2}
\end{figure*}

\begin{figure*}[h]
    \centering
    \includegraphics[width=\linewidth]{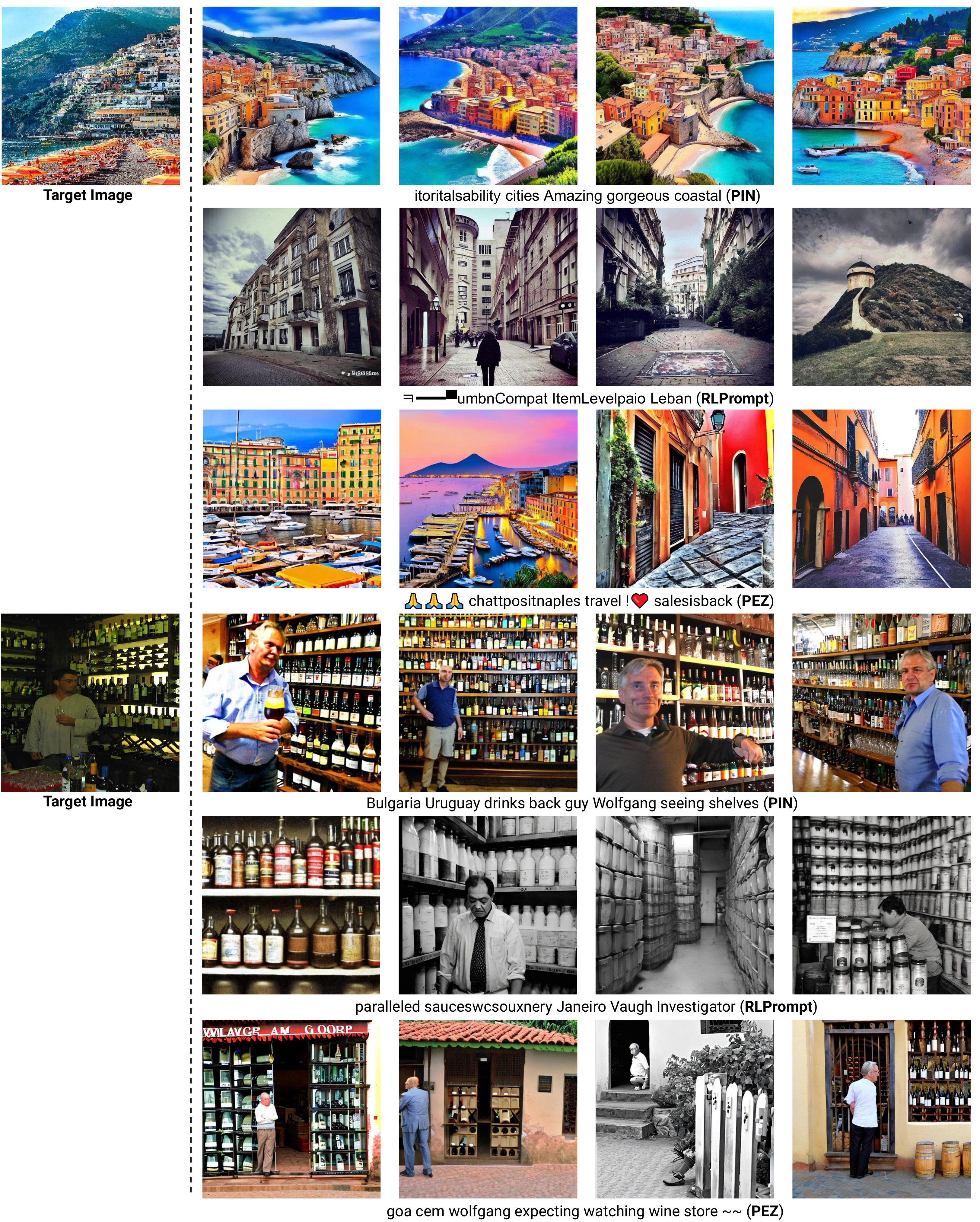}
    \caption{Generated images using learned hard prompts (bottom) through Stable Diffusion-v2~\cite{latent-diff} for a given target image (left).}
    \label{fig:clip_ex_v3}
\end{figure*}

We provide the learned hard prompts by PIN and baselines for few-shot text classification task in Table~\ref{tab:fsc_prompts}.
In textual inversion from images tasks, to visually demonstrate the impact of learned hard prompts, 
we have generated images based on them using a CLIP-based model. Figures~\ref{fig:clip_ex_v1}, ~\ref{fig:clip_ex_v2}, and ~\ref{fig:clip_ex_v3} showcase these images.


\section{Ethics Statement}
We are aware of the potential for misuse, 
particularly in scenarios where the algorithm could be manipulated to favor the generation of biased or toxic content 
by aligning rewards with the toxicity level of the output.


\begin{table*}[t]
\centering
\begin{small}
\resizebox{\linewidth}{!}{
\begin{tabular}{rl}
\toprule
Dataset & SST-2 \\
Instruction & In this task, you are given sentences from movie reviews. 
The task is to classify a sentence as ``great" if the sentiment \\ 
& of the sentence is positive or as ``terrible" 
if the sentiment of the sentence is negative. \\
Manual prompt & [$\boldsymbol{x}$] It was [MASK]. \\
RLPrompt & [$\boldsymbol{x}$] o overall downright just downright [MASK]. \\
PEZ & [$\boldsymbol{x}$] positive positive!)<s> [MASK]. \\
PIN & [$\boldsymbol{x}$] This language delivery feels consistently [MASK]. \\
\midrule
Dataset & Yelp P. \\
Instruction &  In this task, you are given Yelp reviews. 
The task is to classify a review as ``great" if the overall sentiment 
of \\ &the review is positive or as ``terrible" if the overall sentiment of the review is negative.
\\
Manual prompt & [$\boldsymbol{x}$] It was [MASK]. \\
RLPrompt & [$\boldsymbol{x}$] thoroughly… Absolutely downright Absolutely [MASK]. \\
PEZ & [$\boldsymbol{x}$] He collection murderous big Faculty [MASK]. \\
PIN & [$\boldsymbol{x}$] Overall absolutely utter complete absolutely [MASK]. \\
\midrule
Dataset & MR \\
Instruction &  In this task, you are given sentences from movie reviews. The task is to classify a sentence as ``great" if the sentiment \\ & of the sentence is positive or as ``terrible" if the sentiment of the sentence is negative.
\\
Manual prompt & [$\boldsymbol{x}$] It was [MASK]. \\
RLPrompt & [$\boldsymbol{x}$] y overall downright just generally [MASK] \\
PEZ & [$\boldsymbol{x}$] <s>positive positive pharmac restores [MASK]. \\
PIN & [$\boldsymbol{x}$] Total grade absolutely utterly totally [MASK]. \\
\midrule
Dataset & CR \\Instruction &  In this task, you are given sentences from customer reviews. The task is to classify a sentence as ``great" 
\\ & if the sentiment of the sentence is positive or as ``terrible" if the sentiment of the sentence is negative.
\\
Manual prompt & [$\boldsymbol{x}$] It was [MASK]. \\
RLPrompt & [$\boldsymbol{x}$] e pretty downright just downright [MASK]. \\
PEZ & [$\boldsymbol{x}$] <s>immigrant positive and<s> [MASK]. \\
PIN & [$\boldsymbol{x}$] Y word its feeling completely [MASK]. \\
\midrule
Dataset & AG's News \\
Instruction & In this task, you are given a news article. Your task is to classify the article to one out of the four topics
``World", ``Sports", \\ & ``Business", ``Tech" if the article"s main topic is relevant to the world, sports, business,
and technology, correspondingly. 
\\ & If you are not sure about the topic, choose the closest option.
\\
Manual prompt & [MASK] News: [$\boldsymbol{x}$] \\
RLPrompt &  [MASK] … Mum about V [$\boldsymbol{x}$] \\
PEZ & [$\boldsymbol{x}$] Sportsbusiness technology VERY<s> Politics Sports [MASK]. \\
PIN & [MASK] news Ed Sherman Staff Interview [$\boldsymbol{x}$] \\
\midrule
Dataset & Yahoo \\
Instruction & You are given a passage. Using the information present in the passage, you need to classify it into one of 
the 10 topics: 
\\ & 0 - Culture, 1 - Science, 2 - Health, 3 - Education, 4 - Computers, 5 - Sports, 6 - Business, 7
- Music, 8 - Family, 9 - Politics.
\\
Manual prompt & Topic [MASK]: [$\boldsymbol{x}$]. \\
RLPrompt & [$\boldsymbol{x}$] ever people nowadays why some [MASK] \\
PEZ & [$\boldsymbol{x}$] grow neurological the cricket Swift [MASK] \\
PIN & [$\boldsymbol{x}$] to under On Most About [MASK] \\
\midrule
Dataset & Subj \\
Instruction & In this task, you are given sentences from reviews. The task is to classify a sentence as ``subjective" 
\\ & if the
opinion of the sentence is subjective or as ``objective" if the opinion of the sentence is objective.
\\
Manual prompt & [$\boldsymbol{x}$] It was [MASK]. \\
RLPrompt & [$\boldsymbol{x}$] quickly ultimately already four immediately [MASK] \\
PEZ & [$\boldsymbol{x}$] <s> organizationally crimson contexts [MASK]. \\
PIN & [$\boldsymbol{x}$] Daniel is however already is [MASK] \\
\bottomrule
\end{tabular}
}
\end{small}
\caption{Hard prompts that learned by RLPrompt, PEZ, PIN, and Manual instructions following natural instructions~\cite{natural_instruction}.}
\label{tab:fsc_prompts}
\end{table*}

\newpage

\label{sec:appendix}

\end{document}